\pdfoutput=1
\documentclass[11pt]{article}

\usepackage[]{ACL2023}

\usepackage{times}
\usepackage{latexsym}

\usepackage[T1]{fontenc}
\usepackage[utf8]{inputenc}

\usepackage{microtype}

\usepackage{inconsolata}

\usepackage{color}

\usepackage{graphicx}
\usepackage{multirow}

\usepackage{algorithm}
\usepackage{algpseudocode}
\usepackage{multirow}
\usepackage{subcaption}
\usepackage{booktabs}
\usepackage{MnSymbol}
\usepackage{graphicx}
\usepackage{enumitem}
\usepackage{amsmath}

\newcommand{\exampleone}{The pizza was great , but the service was terrible.}

\title{A Weak Supervision Approach for Few-Shot Aspect Based Sentiment Analysis}

\author{Robert Vacareanu$^{2,3}$\thanks{~~Work done during internship at AWS AI Labs} ~~~~Siddharth Varia$^1$  ~~~~ Kishaloy Halder$^1$ ~~~~ Shuai Wang$^1$\\ \textbf{Giovanni Paolini$^1$} ~~~~ \textbf{Neha Anna John$^1$}  ~~~~ \textbf{Miguel Ballesteros$^1$} ~~~~ \textbf{Smaranda Muresan$^{1}$}
 \\  $^1$AWS AI Labs \\ $^2$University of Arizona, Tucson, AZ, USA\\ $^3$Technical University of Cluj-Napoca, Romania \\
 {\tt \footnotesize{\{siddhvar, kishaloh, wshui, paoling, nehajohn, ballemig, smaranm\}@amazon.com}}
\\ {\tt \footnotesize{rvacareanu@arizona.edu}} 
}

\begin{document}
\maketitle
\begin{abstract}
We explore how weak supervision on abundant unlabeled data can be leveraged to improve few-shot performance in aspect-based sentiment analysis (ABSA) tasks.
We propose a 
pipeline approach to construct a noisy ABSA dataset, and we use it to 
adapt a pre-trained sequence-to-sequence model to the ABSA tasks.
We test the resulting model on three widely used ABSA datasets, before and after fine-tuning.
Our proposed method preserves the full fine-tuning performance while showing significant improvements ($15.84\%$ absolute F1) in the few-shot learning scenario
for the harder tasks. In zero-shot (i.e., without fine-tuning), our method outperforms 
the previous state of the art on the aspect extraction sentiment classification (AESC) task and is, additionally, capable of performing the harder aspect sentiment triplet extraction (ASTE) task.
\end{abstract}

\section{Introduction}
\label{s:introduction}
Aspect Based Sentiment Analysis (ABSA) is a fine-grained variant of sentiment analysis \cite{Hu2004MiningAS,pontiki-etal-2014-semeval,pontiki-etal-2015-semeval,pontiki-etal-2016-semeval,Zhang2021AspectSQ,Shu2022ZeroShotAS,Zhang2022ASO},
where the task is to predict the sentiment expressed towards an entity or a certain aspect of an entity, instead of just the sentence-level sentiment (\textit{e.g.,} traditional sentiment analysis tasks \cite{Socher2013RecursiveDM,Santos2014DeepCN}).

For illustration, for a review \textit{The pizza was great, but the service was terrible}, a sentence-level sentiment analysis model might identify the sentiment as \textit{neutral}. The need for ABSA stems from such complex interactions between the target and the polarity of the sentiment 
\cite{pontiki-etal-2014-semeval}.
An ABSA model has to identify the sentiment towards \textit{pizza} as \textit{positive}, and \textit{service} as \textit{negative}, for a holistic understanding of the text.
Furthermore, ABSA tasks can include the identification of the opinion terms (i.e. \textit{great}, \textit{terrible}), and the aspect categories (i.e. \texttt{FOOD},  \texttt{SERVICE})
\cite{Zhang2021AspectSQ}. 

Although traditionally considered as a structured prediction task in the ABSA literature, recent works have shown how sequence-to-sequence (seq-to-seq) models can be effective in these tasks with a generative approach \cite{Yan2021AUG, Zhang2021AspectSQ}. Such approaches leverage the knowledge gained from one task to seamlessly perform well in another. As such, we build upon the Instruction Tuning with Multi-Task Learning approach \cite{Varia2022ABSAIT} and address the following five ABSA tasks:
(i) Aspect-term Extraction (AE), 
(ii) Aspect-term Extraction and Sentiment Classification (AESC), 
(iii) Target Aspect Sentiment Detection (TASD), 
(iv) Aspect Sentiment Triplet Extraction (ASTE), and
(v) Aspect Sentiment Quadruple Prediction (ASQP). 

Sentence-level sentiment annotations are comparatively cheaper and are available at scale through automated proxies (\textit{e.g.,} $\bigstar \text{ or } \bigstar\bigstar$ become \textit{negative}, $\bigstar\bigstar\bigstar\bigstar \text{ or } \bigstar\bigstar\bigstar\bigstar\bigstar$ become \textit{positive}, in the Amazon/Yelp review corpus \cite{Zhang2015CharacterlevelCN}). On the contrary, ABSA requires understanding at sub-sentence level with multiple words or phrases being related to each other, making it prohibitively costly to annotate at scale.\footnote{This is evident from the corpus size of 2.1k vs 700k for \textsc{Rest16} and \textsc{Yelp-full}, respectively.}
However, the abundance of generic review data presents a promising opportunity to improve the performance of a pre-trained language model (PLM) beyond simply fine-tuning it on the small annotated ABSA corpora.

Towards this end, we first construct a noisily annotated ABSA corpus out of generic customer review data without any direct supervision.
We utilize this noisy corpus to pre-train a seq-to-seq model on multiple ABSA tasks. We show that such models are capable of learning in zero/few-shot in final downstream ABSA tasks.
Our contributions are the following: 
(i) We propose a weakly supervised method to obtain annotations for three out of the five ABSA tasks explored in the literature;
(ii) We introduce a pre-training step to improve the few-shot performance on the downstream task of PLMs; 
(iii) We comprehensively evaluate our proposed method in three scenarios (full fine-tuning, few-shot, and zero-shot learning), yielding as much as $15.84\%$~F1 improvement over the SOTA baselines. We release the sources to create the few-shot benchmarking datasets\footnote{\url{https://github.com/robertvacareanu/NoisyABSAPreTraining}}.

\section{Related Work}
\label{s:related_work}

Aspect-Based Sentiment Analysis has received tremendous attention in the past years \cite{Tulkens2020EmbarrassinglySU,Zhang2021AspectSQ,Shu2022ZeroShotAS,Zhang2022ASO}, 
either handling single tasks, such as aspect term extraction \cite{He2017AnUN,Liu2015FinegrainedOM,Tulkens2020EmbarrassinglySU}, aspect category detection \cite{Tulkens2020EmbarrassinglySU}, 
aspect sentiment classification \cite{Vo2015TargetDependentTS,Xu2019BERTPF,Li2021DualGC,Wang2021EliminatingSB}, or handling compound tasks \cite{Zhang2015NeuralNF, Yu2021SelfQA, Xu2020PositionAwareTF, Zhang2021AspectSQ}.
For the latter group, it typically includes either a pipeline approach \cite{Peng2020KnowingWH,Yan2021AUG} or an end-to-end (E2E) approach \cite{Xu2020PositionAwareTF,Zhang2021AspectSQ,Zhang2021TowardsGA}.

In the pipeline approach the final prediction is constructed using the output of multiple components. The disadvantage of such models is that the error is propagated throughout the system \cite{Zhang2022ASO}.

In the E2E approach, the model learns the interactions jointly between the multiple prediction tasks, which is believed to improve the final performance \cite{Xu2020PositionAwareTF,Zhang2022ASO}. Our proposed approach falls in this category.
Typical E2E approaches include: (i) treating it as a token classification task \cite{Xu2019BERTPF,Shu2019ControlledCS,Xu2020PositionAwareTF}, 
(ii) framing it as a machine reading comprehension task \cite{Chen2021BidirectionalMR,Liu2022ARO}, natural language inference task \cite{Shu2022ZeroShotAS},
or as a language generation task \cite{Zhang2021TowardsGA,Yan2021AUG,Zhang2021AspectSQ,Varia2022ABSAIT}. 

Our proposed approach treats the ABSA tasks as a generation task, similar to \cite{Zhang2021AspectSQ,Varia2022ABSAIT}. 
We build upon the paradigm called Instruction Tuning with in Multi-Task Learning (IT-MTL), introduced in \cite{Varia2022ABSAIT}, resulting in a single model capable of handling different ABSA tasks. 
However, none of these methods takes advantage of the vast amount of review data available, other than just pre-training on them with some generic language modeling objectives.
\section{Method}
\label{s:method}
We introduce an additional step in the classical \texttt{pretrain} $\rightarrow$ \texttt{finetune} approach \cite{Howard2018UniversalLM, Devlin2019BERTPO, Raffel2020ExploringTL}, transforming it into \texttt{pretrain} $\rightarrow$ \texttt{\textbf{N}oisy \textbf{A}BSA \textbf{P}re-\textbf{T}raining (NAPT)} $\rightarrow$ \texttt{finetune} for ABSA.
We propose an approach for building a weakly annotated dataset for the intermediate NAPT step. We use this noisy dataset to enhance the knowledge of a pretrained model with the intuition that exposing the model to tasks which are well aligned with the final downstream task, improves the performance. We then consider this as the backbone base model, and finetune it on the downstream task as usual.
Our proposed approach is applicable to any generic seq-to-seq model.

\subsection{Dataset Construction}
The first step in our proposed method is to weakly annotated a dataset
without any direct supervision.\footnote{We use models which were trained on different tasks, but no model has seen any aspect-based sentiment analysis data.} Our proposed approach annotates a dataset with tuples of the form aspect-terms, opinion-terms, and sentiment polarity. We follow a pipeline approach as shown in Table \ref{tab:pipeline}\cite{xu-etal-2013-mining, Zhang2022ASO}, but without using any direct ABSA supervision. We describe each step in greater detail next.

\begin{table*}[]
\centering
\resizebox{0.8\textwidth}{!}{%
\begin{tabular}{|ccc|}
\hline
\multicolumn{3}{|c|}{\textbf{Sentence:} \textit{The pizza was great, but the service was terrible.}} \\ \hline
\multicolumn{1}{|c|}{\textbf{Step}} & \multicolumn{1}{c|}{\textbf{Heuristic / Method}} & \textbf{Resulting Annotations} \\ \hline
\multicolumn{1}{|c|}{\#1} & \multicolumn{1}{c|}{Extract frequent nouns as Aspect-terms} & pizza, service \\ \hline
\multicolumn{1}{|c|}{\#2} & \multicolumn{1}{c|}{Extract matches with an opinion lexicon as Opinion-terms} & great, terrible \\ \hline
\multicolumn{1}{|c|}{\#3} & \multicolumn{1}{c|}{\begin{tabular}[c]{@{}c@{}}Predict entailment of form \textit{\{aspect\} is \{opinion\}} for every \\ aspect, opinion combinations using a pre-trained NLI model\end{tabular}} & \textless{}pizza, great\textgreater{}, \textless{}service, terrible\textgreater{} \\ \hline
\multicolumn{1}{|c|}{\#4} & \multicolumn{1}{c|}{\begin{tabular}[c]{@{}c@{}}Classify documents of form \textit{\{aspect\} is \{opinion\}} with a \\ pre-trained sentiment analysis model\end{tabular}} & \begin{tabular}[c]{@{}c@{}}\textless{}pizza, great, positive\textgreater{},\\ \textless{}service, terrible, negative\textgreater{}\end{tabular} \\ \hline
\end{tabular}%
}
\caption{A step-by-step illustration of our noisy dataset construction pipeline. It follows a pipeline approach, and yields <aspect, opinion, sentiment> triplets in the end for each sentence in a generic review corpus.}
\label{tab:pipeline}
\end{table*}

\subsubsection{Aspect-term Extraction}
\label{sss:ate}

The first step in our proposed dataset creation procedure is aspect-term extraction. 
We use {\tt spacy} tokenizer to obtain POS tags and then consider $20\%$ of the most frequent nouns in the text. These nouns serve as candidate aspect terms.
We note that this method implicitly assumes that dataset $D$ consists of a single domain. Nevertheless, this is a small assumption as the reviews are typically directed towards a product of a known category \cite{He2016UpsAD, Zhang2015CharacterlevelCN}.
We extend this method to multi-word aspect terms by considering collocations of length $\le$ 4 filtered by their POS tags. For example, we allow bigrams of the form 
{\tt NN-NN} like {\it chicken breast} (\textit{cf} Table \ref{appendix_ss:multiword_patterns} for all patterns used).
Finally, we filter out the sentences from which no aspect term was extracted.

\subsubsection{Opinion-term Extraction}
\label{sss:ote}

The second step in our proposed algorithm is opinion term extraction. We take a lexicon-based approach to opinion extraction \cite{Ding2008AHL, Kanayama2006FullyAL, Hu2004MiningAS}. In particular, we use the opinion lexicon from \cite{Hu2004MiningAS} and perform word matching on the target text. If negations \textit{e.g.,} \textit{no} or \textit{not} appear before the opinion word, we include it in the final extraction as well. 
We filter out the sentences from which no opinion term was extracted.

\subsubsection{Linking Opinion-terms with Aspect-terms}
\label{sss:ate_ote_link}

So far the resulting dataset consists of noisy aspect, and opinion terms, but without the association between them. For example, for a sentence such as \textit{\exampleone}, the proposed algorithm would extract \textit{pizza} and \textit{service} as the aspect terms and \textit{great} and \textit{terrible} as the opinion terms, respectively. But at this point we do not know that \textit{great} refers to \textit{pizza} and \textit{terrible} refers to \textit{service}. 
We reformulate this problem as a natural language inference problem \cite{Dagan2005ThePR, Shu2022ZeroShotAS}. We use an {\tt MPNet}\footnote{huggingface.co/symanto/mpnet-base-snli-mnli} model \cite{Song2020MPNetMA} and construct artificial sentences to determine which opinion-term refers to which aspect-term.
More precisely, we construct sentences such as \texttt{<aspect-term> is <opinion-term>}, for each aspect- and opinion-term.\footnote{We
relax strict grammatical correctness \textit{e.g.,} the formulation might result in 
\textit{burgers is great} instead of \textit{burgers are great}).}
Then, we use the original sentence (e.g. \textit{\exampleone}) as the premise and our artificially constructed sentence as the hypothesis (e.g. \textit{pizza is great}). We interpret a high entailment score ($\ge0.75$) as evidence that the opinion term refers to that particular aspect term.
We discard aspect- and opinion-term pairs where the entailment score was below the threshold.

\noindent{\textbf{Alternative Approach:} We consider %constituency-based 
an alternate approach where the linking is based on 
constituency-parse rules which turns out disadvantageous. Constituency parsing is considerably slower and the rules are non-trivial to formulate.}

\subsubsection{Sentiment Extraction}
\label{sss:sentiment}

The last step in our proposed dataset creation method is to add the sentiment \cite{Hu2004MiningAS} to each {\tt <aspect-term, opinion-term>} tuple. We use a sentence-level classifier on top of artificially constructed sentences \cite{Sanh2019DistilBERTAD}. For example, for a tuple such as <pizza, great>, we feed the sentence \textit{pizza is great} through a sentence-level sentiment classifier.\footnote{huggingface.co/distilbert-base-uncased-finetuned-sst-2-english} Then, we label the <aspect term, opinion term> tuple with the sentiment prediction if the model's confidence is above a certain threshold ($\ge0.75$), otherwise we discard the tuple.
At the end of this step, for the sentence \textit{\exampleone} we have the following {\tt <aspect-term, opinion-term, sentiment>} noisy annotations: <pizza, great, positive>, <service, terrible, negative>. We consider an alternative for this step using the sentiments associated in the opinion lexicon, but a classifier allows for confidence filtering.

Throughout our proposed dataset creation process we use external resources, such an opinion lexicon, an NLI model and a sentence-level sentiment classifier. However, these resources do not consume any annotated ABSA data by any means.

\subsection{Noisy ABSA Pre-training (NAPT)}
\begin{figure*}[t]
    \centering
    \includegraphics[width=0.9\textwidth]{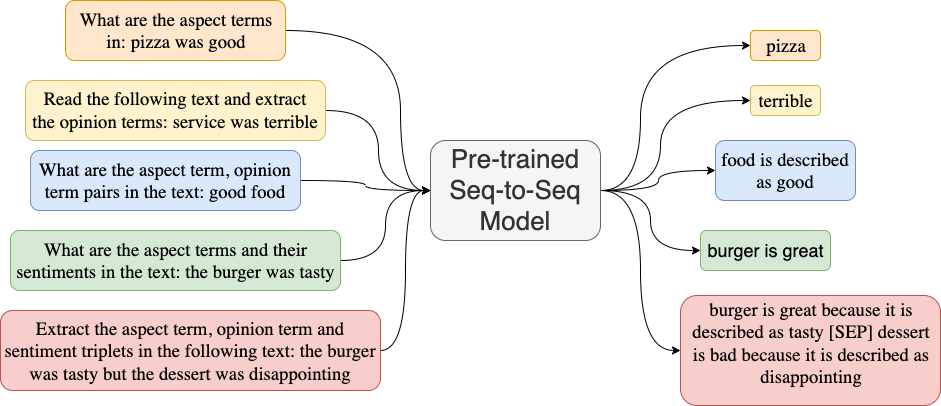}
    \caption{Overview of our proposed Noisy ABSA Pre-Training (NAPT). We start from a pretrained language model and extend its capabilities by instruction tuning it in a multi-task learning fashion. We use $5$ different yet related tasks for the proposed NAPT step. The tasks we use are: (i) aspect-term extraction, (ii) opinion-term extraction, (iii) aspect-term extraction and opinion-term extraction, (iv) aspect term extraction and sentiment classification, and (v) aspect-term extraction, opinion-term extraction, and sentiment classification. This step results in a model capable of performing multiple ABSA tasks.}
    \label{fig:pre_finetuning_step}
    \vspace{-5mm}
\end{figure*}

The phase consists of exposing the model to tasks that are more aligned with the final downstream task, 
\textit{i.e.,} ABSA in our case. We factorize the triplets from the noisy dataset into five separate but overlapping tasks: (i) aspect-term extraction, (ii) opinion-term extraction, (iii) aspect-term and opinion-term extraction, (iv) aspect-term extraction and sentiment prediction, and (v) aspect-term extraction, opinion-term extraction and sentiment prediction. Note that there exists a correspondence between our NAPT tasks and classical ABSA tasks: tasks (i), (iv) and (v) correspond to Aspect Extraction (AE), Aspect Extraction Sentiment Classification (AESC), and Aspect Sentiment Triplet Extraction (ASTE), respectively. We use the noisy ABSA dataset to pre-train the base model. We train the model parameters in a multi-task learning framework (\textit{cf} Figure \ref{fig:pre_finetuning_step}) using instruction tuning with a diverse set of instructions \cite{Sanh2022MultitaskPT}. At the end of NAPT, the resulting model is imbued with the capability of performing multiple ABSA tasks. This can serve as a drop-in replacement to the off-the-shelf pre-trained checkpoints that are widely used in the generative ABSA literature.

\subsubsection{Addressing Overfitting}
The primary goal of our proposed NAPT phase is to \textit{enhance} the pre-trained model while retaining existing knowledge from pre-training objectives, in other words, avoiding catastrophic forgetting and overfitting. We achieve this in a few different ways. First, instead of just randomly splitting the data into train/validation, we split the extracted aspect- and opinion-terms into two disjoint sets, favoring novel aspect- and opinion term constructions in the validation partition. 
We observe this split definition to be necessary to prevent overfitting of the base model.
Additionally, we invoke three types of regularization:
\begin{itemize}[leftmargin=*]
    \item \textbf{Standard weight decay:} we add a standard $\ell^2$ regularization term to the loss function.
    \item \textbf{Tuple Dropout:} we apply dropout over the tuples that the model is trained to extract to prevent it from overfitting to the noisy annotations. We randomly dropped $50\%$ of the tuples from prediction targets of the seq-to-seq model.
    \item \textbf{Biased weight decay:} we use a biased variant of weight decay to prevent the parameters from diverging considerably from the initialization point, akin to \cite{Kirkpatrick2017OvercomingCF}. Towards this, we use the $\ell^2$ norm over the difference between the current ($\theta$) and the initial weights of the model ($\theta_{init}$), and add it to the loss.\\ 
    Our final loss function ($\mathcal{L}$) is:
    \begin{equation}
        \mathcal{L} = CE_{loss} + \alpha \cdot \ell^2(\theta-\theta_{init}) +\beta \cdot \ell^2(\theta).
        \label{eq:loss}
    \end{equation}
    where $\alpha$ and $\beta$ are hyperparameters, and $CE_{loss}$ denotes the standard cross-entropy loss.
\end{itemize}

\section{Experiments}
\label{s:experiments}

We compare against state-of-the-art methods on three widely used ABSA datasets. We evaluate in three scenarios: 
(i) $k$-shot learning: where the model has access to at least $k$ examples of each 
class, 
(ii) zero-shot evaluation: where the model has not seen any example at all from the gold-annotated ABSA data,
and (iii) full-training: where the model has access to the complete gold-standard training data,

\subsection{Experimental Setup}
In all our experiments, we use \texttt{T5} \cite{Raffel2020ExploringTL}, particularly \texttt{t5-base} as the pre-trained seq-to-sed model, which has $\sim220$M parameters. 
We experiment with \texttt{t5-large} as well to explore the impact of model size on the downstream performance (\textit{cf} Appendix \ref{appendix:all_experiments}).
We use the standard evaluation metrics as previous work, which is F1 score over the exact match of the tuples. 
For zero-shot, we use the same evaluation procedure as \cite{Shu2022ZeroShotAS}, which is token-level F1 score.

We use a random subset of \textit{Amazon Electronics} \cite{He2016UpsAD}, and \textit{Yelp} reviews \cite{Zhang2015CharacterlevelCN} to 
create our noisy-annotated dataset.\footnote{100K reviews from Amazon, and YELP each are used.} 
We split the reviews with $\ge 3$ sentences using a sentence tokenizer.
We split the noisy dataset into train/validation split. 
We enforce that 
there is no overlap in terms of aspect-terms between the train/validation splits.
This results in approximately $190$k examples for training and $12.5$k examples for validation.

We repeat each experiment with $5$ different random seeds. Additionally, we repeat the noisy ABSA pre-training step with $3$ different random seeds. 
As a result, the numbers corresponding to our proposed method (i.e. the ones with \texttt{-APT}) represent an average of $5\times3=15$ runs, and all the other numbers represent an average of $5$ runs. We report the mean and (sample) standard deviation. 

We present the results on the Aspect Sentiment Triplet Extraction (ASTE) and Aspect-term Extraction and Sentiment Classification (AESC) tasks available in all the datasets we use for evaluation.\footnote{Results for \textbf{all} tasks are in Tables \ref{appendix_ss:kshot_lap14},\ref{appendix_ss:kshot_rest15},\ref{appendix_ss:kshot_rest16}, and \ref{appendix_ss:full_training_lap14},\ref{appendix_ss:full_training_rest15},\ref{appendix_ss:full_training_rest16} for $k$-shot and full training respectively.}

\subsection{Datasets}
\label{ss:datasets}
We use three popular datasets for aspect-based sentiment analysis: 
\textsc{Rest15}, \textsc{Rest16} and \textsc{Lap14} \cite{pontiki-etal-2014-semeval, pontiki-etal-2015-semeval, pontiki-etal-2016-semeval}, 
which cover two domains: restaurant and laptop, respectively. 
In particular, we use the version released by \citeauthor{Zhang2021AspectSQ}. 
For $k$-shot, we use the same splits as \cite{Varia2022ABSAIT} to ensure a fair comparison. 
Specifically, the k-shot datasets were created by sampling $k$ examples for each attribute. 
The attributes are \textit{aspect category}, and \textit{sentiment} for restaurant, and laptop respectively.

\subsection{Baselines}
\label{ss:baselines}
Since we introduce the NAPT step and build upon the existing Instruction Tuning with Multi-Task Learning ({\tt IT-MTL}) paradigm, we refer to our proposed method as {\tt IT-MTL-NAPT}. We compare this with standard fine-tuning based approaches that generally show strong performance in ABSA tasks \textit{i.e.,}
(i) text-only (\texttt{Text}), where we give the model the text review and train it to predict the gold text \cite{Zhang2021AspectSQ}, 
(ii) instruction tuning (\texttt{IT}) and 
(iii) instruction tuning + multi-task learning, as per \cite{Varia2022ABSAIT} (\texttt{IT-MTL}).

To succinctly show the effectiveness of proposed NAPT, we keep another baseline where a seq-to-seq model is further pre-trained with in-domain data using the same objective as that of {\tt t5} \textit{i.e.,} span prediction. We call it \texttt{IT-MTL-ID}.\footnote{As in, In-Domain (ID) pre-training occurs along with IT-MTL.} The in-domain data is essentially the same as that of the NAPT corpus, but without the noisy annotations.

\begin{figure*}[!t]

    \begin{subfigure}[b]{0.31\textwidth}
        % \centering
        \includegraphics[width=1.1\columnwidth]{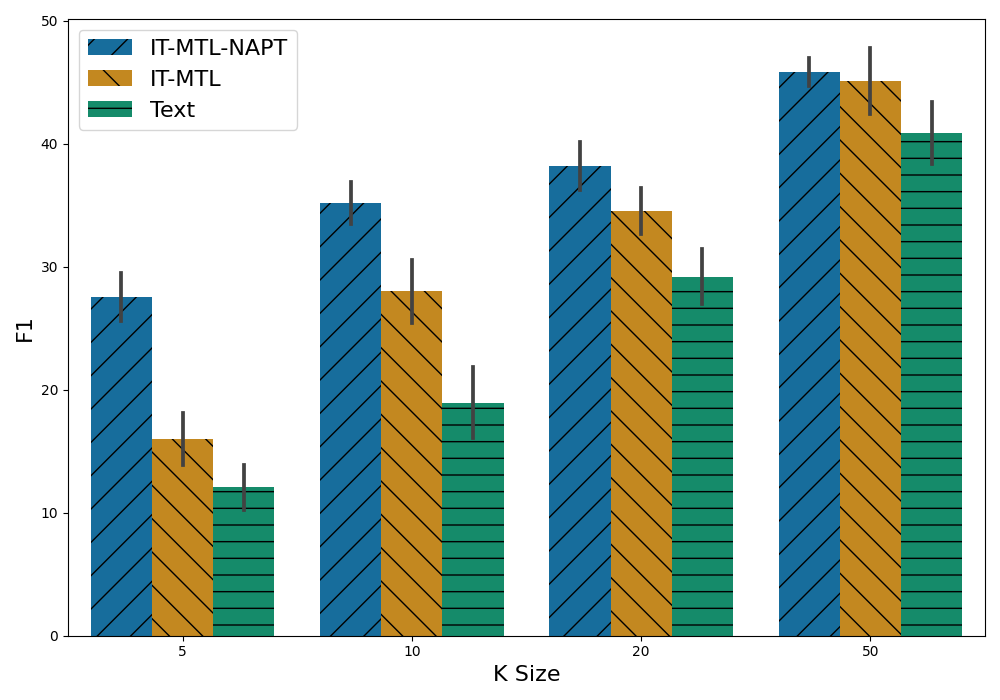}
        \caption{\footnotesize{\textsc{Lap14} on ASTE Task}}
        \label{fig:lap14_kshot}
    \end{subfigure}
    \hfill
    \begin{subfigure}[b]{0.31\textwidth}
        % \centering
        \includegraphics[width=1.1\columnwidth]{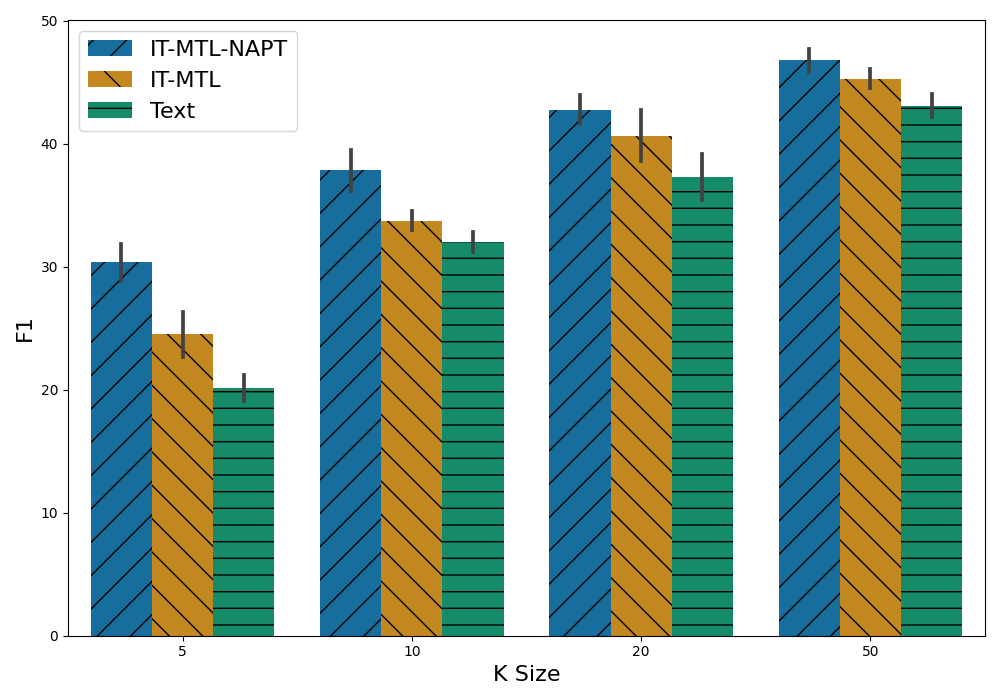}
        \caption{\footnotesize{\textsc{Rest15} on ASTE Task}}
        \label{fig:rest15_kshot}
    \end{subfigure}
    \hfill
    \begin{subfigure}[b]{0.31\textwidth}
        % \centering
        \includegraphics[width=1.1\columnwidth]{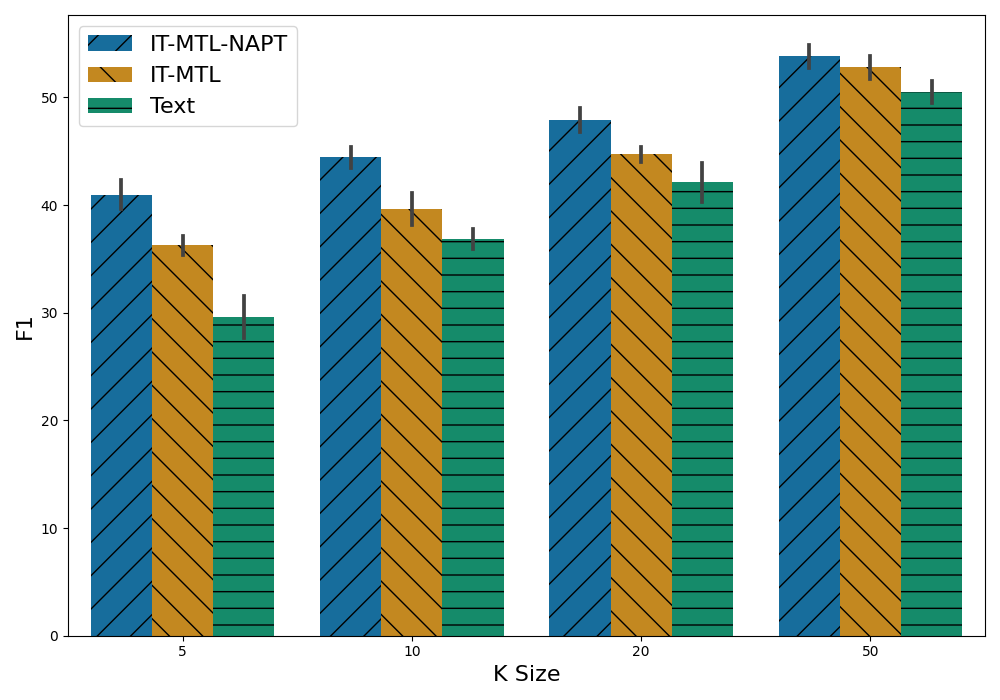}
        \caption{\footnotesize{\textsc{Rest16} on ASTE Task}}
        \label{fig:rest16_kshot}
    \end{subfigure}
    \\
    \begin{subfigure}[b]{0.31\textwidth}
        % \centering
        \includegraphics[width=1.1\columnwidth]{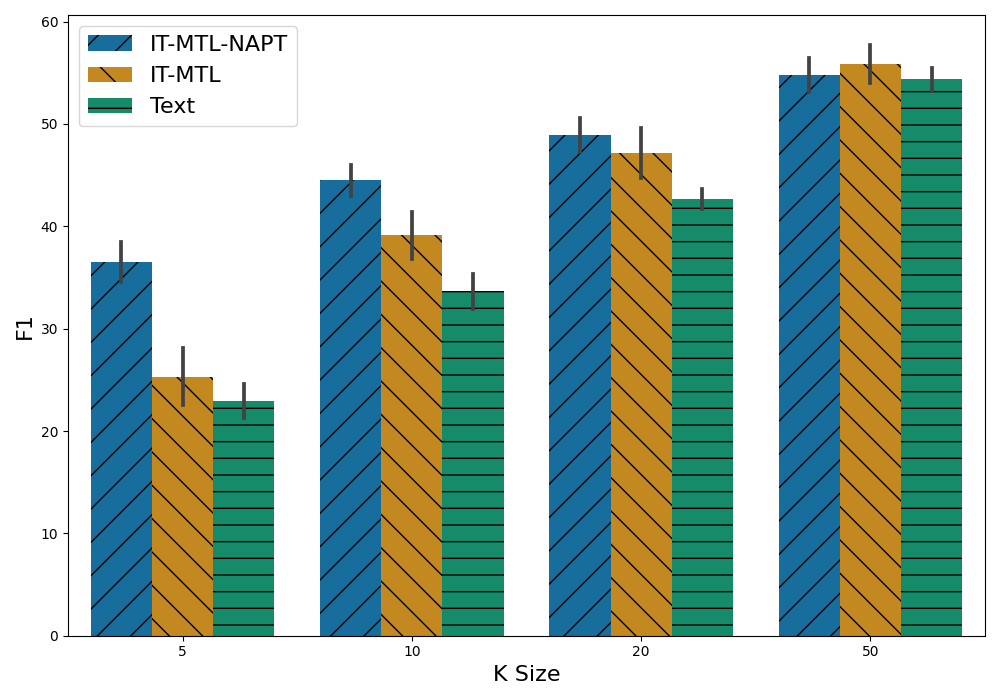}
        \caption{\footnotesize{\textsc{Lap14} on AESC Task}}
        \label{fig:lap14_kshot_task2}
    \end{subfigure}
    \hfill
    \begin{subfigure}[b]{0.31\textwidth}
        % \centering
        \includegraphics[width=1.1\columnwidth]{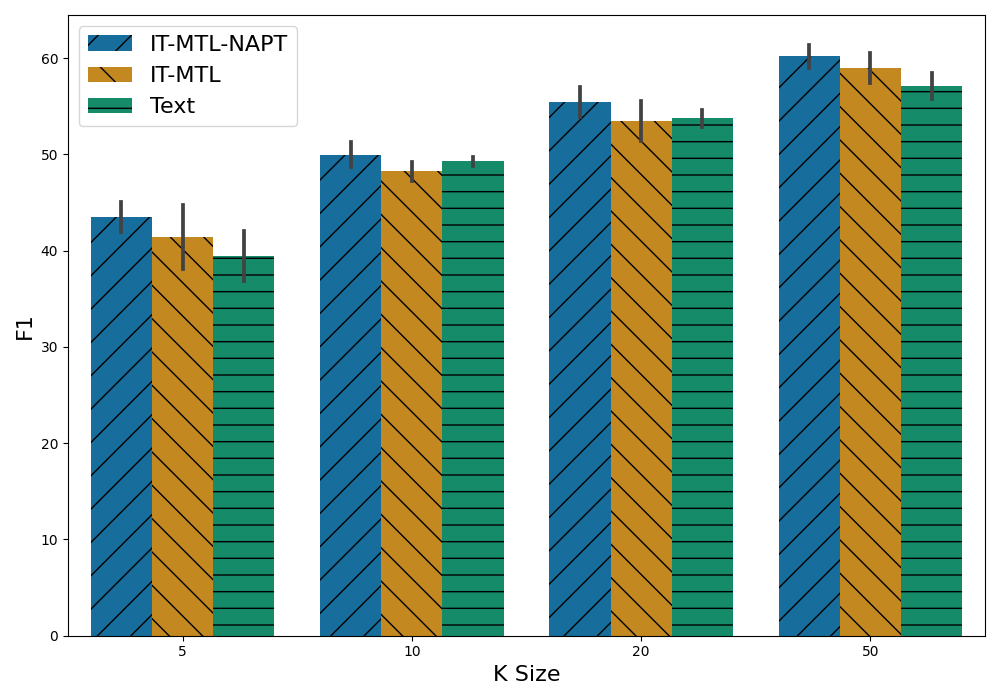}
        \caption{\footnotesize{\textsc{Rest15} on AESC Task}}
        \label{fig:rest15_kshot_task2}
    \end{subfigure}
    \hfill
    \begin{subfigure}[b]{0.31\textwidth}
        % \centering
        \includegraphics[width=1.1\columnwidth]{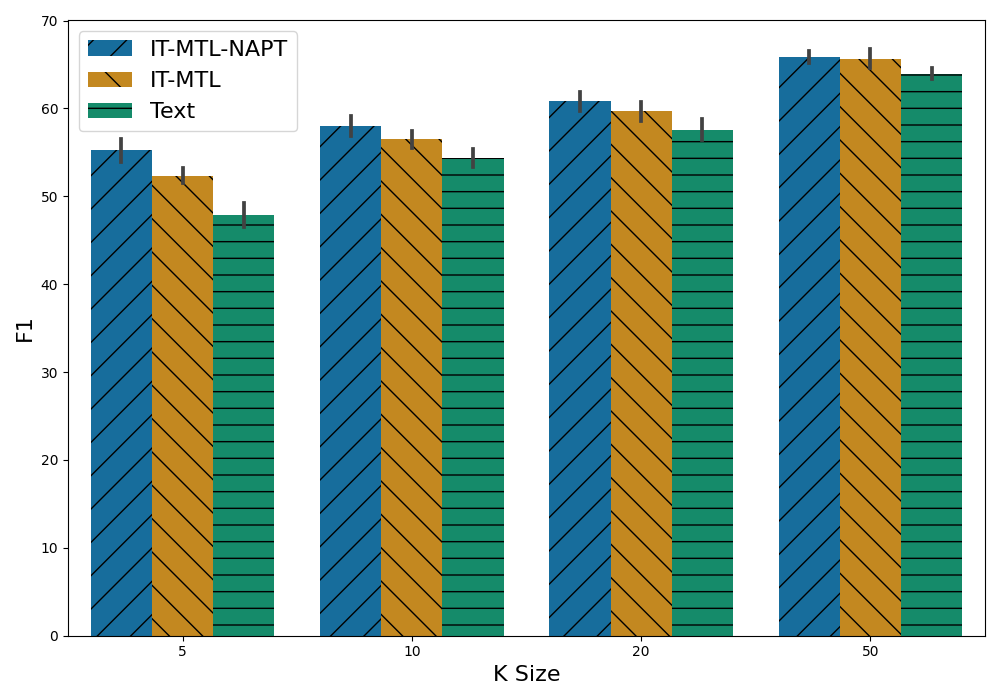}
        \caption{\footnotesize{\textsc{Rest16} on AESC Task}}
        \label{fig:rest16_kshot_task2}
    \end{subfigure}

    \caption{
        Performance Comparison between our proposed method (\texttt{IT-MTL-NAPT}) and two baselines 
        over 3 datasets on on the Aspect Sentiment Triplet Extraction (ASTE), Aspect-term Extraction and Sentiment Classification (AESC) tasks in top, and bottom rows respectively.
        We note that our proposed method helps in all the k splits. \tiny{(larger is better)}}
    \label{fig:kshot_all}
\end{figure*}

\subsection{K-Shot Learning}
\label{ss:k_shot_learning}
Next, we compare between the two approaches in $k$-shot learning scenarios.
We summarize our results in Figure~\ref{fig:kshot_all}. 
\texttt{IT}, and \texttt{IT-MTL-ID} perform similarly with the other baselines, so we skip them for clarity. We include all our results in Appendix~\ref{appendix_ss:kshot_learning}.
First we observe that, our proposed method outperforms the baselines across all datasets in all $k$-shot scenarios, yielding as much as $15.84\%$ F1 points (i.e. from $13.04\%$F1 to $28.88\%$F1) of improvement. 
Second, the performance improvement increases as the number of examples decrease, with the biggest improvement being in the \texttt{k=5} case. This is expected because with the growing number of examples, all models are able to learn the task better. When using the full dataset, as we see in Table~\ref{table:full_training_performance}, both the proposed model and the baseline performances converge. Additionally, we observe that our proposed method brings the larger improvements on the harder tasks, as it gets difficult for the baselines to learn from only a few of examples.

\subsection{Zero-Shot Evaluation}
\label{ss:zse}
Our proposed NAPT step enables the model to perform the following ABSA tasks in zero-shot
\textit{i.e.,} without any gold-standard supervision: 
(i) Aspect-term Extraction (AE), 
(ii) Aspect-term Extraction and Sentiment Classification (AESC), and 
(iii) Aspect Sentiment Triplet Extraction (ASTE). 
We perform two experiments in the zero-shot setting.
First, we investigate how much data does a baseline need to reach the performance obtained by our proposed model in the zero-shot setting. 
Second, we compare against previous work in the ASTE task \cite{Shu2022ZeroShotAS}. 

\subsubsection{Dataset Size Equivalence}
\label{sss:dataset_size_equivalent}
\begin{figure}[t]
    \centering
    \begin{subfigure}[b]{0.23\textwidth}
        \centering
        \includegraphics[width=\textwidth]{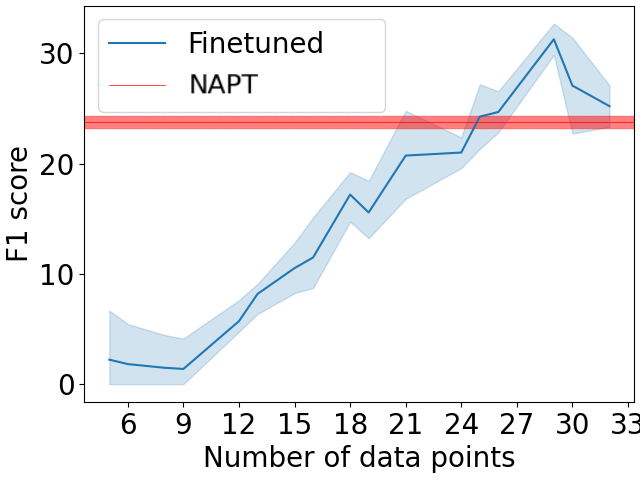}
        \caption{\textsc{Lap14}}% on ASTE Task}
        \label{fig:lap14_0shot}%
    \end{subfigure}
    ~
    \begin{subfigure}[b]{0.23\textwidth}
        \centering
        \includegraphics[width=\textwidth]{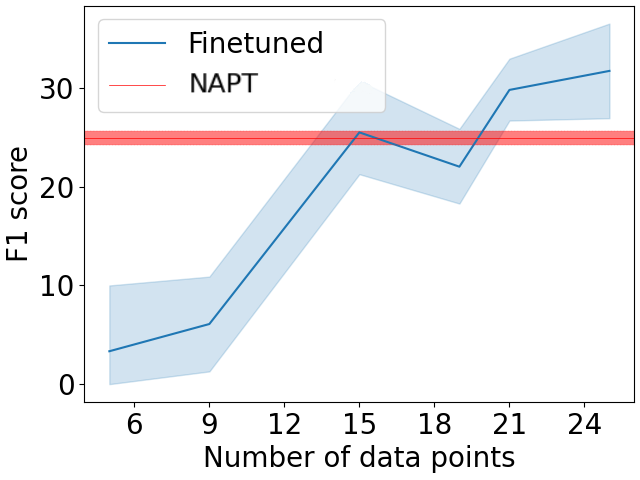}
        \caption{\textsc{Rest15}}% on ASTE Task}
        \label{fig:rest15_0shot}
    \end{subfigure}

    \caption{
        Data size equivalence comparison between {\tt t5} models that are {\tt finetuned} on downstream corpus vs our proposed NAPT for ASTE task in (a) \textsc{Lap14}, (b) \textsc{Rest15} respectively. The finetuned models need $\sim15-25$ completely annotated data points to equalize our proposed method.}
    \label{fig:zs_equivalence}
\end{figure}

We compare our proposed method in zero-shot setting against a baseline model trained on gold-annotated data, where we vary the number of training data points. 
This experiment shows how many annotated data points, on average, is the noisy ABSA pre-training phase equivalent of. 
We observed that the improvement depends on the difficulty of the task and of the dataset, respectively. 
For example, Figure~\ref{fig:zs_equivalence} shows that for the ASTE task, 
one would need $\sim15, 25$ annotated data points to obtain a comparable performance with our proposed method for \textsc{Rest15} and \textsc{Lap14} respectively.
We remark that the number of data points vary according to the difficulty of the task and with the difficulty of the dataset, 
ranging between $\sim6-25$ data points for AE, and ASTE task for \textsc{Lap14} respectively.

\subsubsection{Performance Comparison with Baselines}
\label{sss:zero_shot_performance}
We compare the zero-shot performance of our proposed method with previous work on ABSA \cite{Shu2022ZeroShotAS}, summarized in Table~\ref{table:zsp}. 
Our proposed model outperforms the previous state-of-the-art results for AESC by as much as $6.94\%$F1 points in the
restaurant domain. 
The improvement for the laptop domain is smaller, 
we attribute this to the NAPT dataset being biased towards the restaurant domain in terms of size.
It is interesting to note that our model's backbone \textit{i.e.,} \texttt{t5-base} is able to outperform CORN altough it has almost half the number of parameters as that of its counterpart \textit{i.e.,} \texttt{bart-large}.
\begin{table}[h]
    \centering
    \resizebox{0.7\columnwidth}{!}{
    \begin{tabular}{lcc}
\hline
\textbf{Model} & \textbf{\textsc{Rest}} & \textbf{\textsc{Lap}} \\\hline
CORN & $37.20$ \tiny{$\pm 0.50$} & $40.30$ \tiny{$\pm 0.60$} \\
% IT-MTL-PFT & & \\
% \quad - bart-large & $\mathbf{45.06}$ \tiny{$\pm 0.22$}                                              & $38.70$ \tiny{$\pm 0.46$}                                                      \\
IT-MTL-NAPT & $\mathbf{44.14}$ \tiny{$\pm 0.30$} & $\mathbf{40.51}$ \tiny{$\pm 0.43$} \\ \hline  
\end{tabular}
    }
    \caption{
        Comparison of our proposed method with previous work on zero-shot Aspect Extraction Sentiment Classification (AESC). 
        Our proposed method outperforms the previous work on both datasets. Metric is token-level F1 score.
    }
    \label{table:zsp}
\end{table}

\subsection{Full-Training}
\label{ss:full_training}
We compare the performance of our proposed method (i.e. \texttt{pretrain} $\rightarrow$ \texttt{NAPT} $\rightarrow$ \texttt{finetune}) 
with the standard method of \texttt{pretrain} $\rightarrow$ \texttt{finetune} 
and report the result in Table~\ref{table:full_training_performance}, for all the datasets. 
Overall in the full-training scenario, our proposed method performs comparably with or better than the baseline.
We observe during our preliminary experiments that the training dynamics change drastically between the 
\texttt{pretrain} $\rightarrow$ \texttt{NAPT} $\rightarrow$ \texttt{finetune} and \texttt{pretrain} $\rightarrow$ \texttt{finetune}.

\begin{table}[]
    \centering
    \resizebox{1.0\columnwidth}{!}{
        \begin{tabular}{lrrr}
    \hline
    \multicolumn{1}{c}{\multirow{2}{*}{Model}} & \multicolumn{3}{c}{Dataset}                                                         \\
    \multicolumn{1}{c}{}                       & \multicolumn{1}{l}{\textsc{Lap14}} & \multicolumn{1}{l}{\textsc{Rest15}} & \multicolumn{1}{l}{\textsc{Rest16}} \\ \hline
    Text                                & $59.50\pm 1.35$               & $51.74\pm0.84$                 & $62.95\pm0.61$                 \\
    IT                                  & $\mathbf{60.47} \pm 1.36$     & $52.78\pm0.81$                 & $\mathbf{63.77}\pm0.82$         \\
    IT-MTL                              & $60.17\pm1.19$                & $53.17\pm0.67$                 & $62.69\pm0.69$                 \\
    IT-MTL-ID                           & $58.24\pm1.03$                & $53.42\pm1.27$                 & $62.38\pm0.69$                 \\ \hline
    IT-MTL-NAPT                          & $59.97\pm1.28$                & $\mathbf{53.57}\pm1.42$                 & $61.67\pm0.65$ \\ \hline 
\end{tabular}
    }
    \caption{F1 scores of our proposed method (\texttt{IT-MTL-NAPT}) and $4$ competitive baselines on the Aspect Sentiment Triplet Extraction task over $3$ datasets under training on full dataset. We observe similar levels of performance. 
    }
    \label{table:full_training_performance}
    \vspace{-5mm}
\end{table}

\section{Discussion}
\label{s:discussion}
In this section, we would like to discuss a few important aspects of our approach apart from the main experiments.

\subsection{Ablation}
\label{ss:ablation}
To better understand how different components of our NAPT strategy influence the final downstream performance, 
we conduct the following ablation studies.

\noindent{\textbf{Regarding NAPT Tasks:}} We analyze the importance of NAPT with multiple tasks and their impact on the downstream performance. Our analysis shows that there exists a positive correlation between the NAPT complexity and downstream performance. We average the downstream performance across every task and every $k$-shot split and train on the downstream task in a multi-task learning fashion. 
We summarize our results in Table~\ref{table:ablation_tasks}. 
Our experiments show that it helps in general to align the NAPT and finetuning objectives.
If the NAPT phase is done in a multi-task learning fashion, 
it is beneficial for the model if the same is done for finetuning on the downstream 
task as well. Additionally, we observe that that harder NAPT tasks are beneficial for the downstream task regardless of the way the training on the downstream task is performed, as the F1 scores reflect the relative order in difficulty of the tasks (\textit{i.e.,} ASTE $>$ AESC $>$ AE).

\begin{table}[h]
    \centering
    \resizebox{0.5\textwidth}{!}{
    \begin{tabular}{cccc}
\hline
\multirow{2}{*}{\begin{tabular}[c]{@{}c@{}}\textbf{NAPT}\\ \textbf{Task}\end{tabular}} & \multicolumn{3}{c}{\textbf{Dataset}} \\
 & \textbf{\textsc{Lap14}} & \textbf{\textsc{Rest15}} & \textbf{\textsc{Rest16}} \\ \hline
AE & $43.47$ & $46.72$ & $50.76$          \\
AESC & $44.94$ & $46.99$  & $50.75$       \\
ASTE & $46.30$ & $47.14$ & $51.17$        \\
MTL & $47.45$ & $47.32$ & $51.65$         \\\hline
\end{tabular}

% ablation_data = pd.read_csv('results/ablation_dev_taskwise.tsv', sep=',', index_col=0).drop(columns=['index'])
% ablation_data[((ablation_data['k'] != 'Full Dataset') & (ablation_data['training_style'] == 'mtl'))].groupby(by=['pft_task', 'dataset']).agg({'f1': ['mean']}).unstack(level=['dataset']).round(2).reindex(['ate', 'ate_sent', 'ate_ote_sent', 'mtl']).rename(index={'ate': 'ATE', 'ate_sent': 'AESC', 'ate_ote_sent': 'ASTE', 'mtl': 'MTL'})

% 43.47 46.72 50.76
% 44.94 46.99 50.75
% 46.30 47.14 51.17
% 47.45 47.32 51.65
    }
    \caption{Ablation study over NAPT tasks in terms of macro F1 scores averaged across all the tasks and $4$ $k$-shot settings. It shows that having all the tasks during NAPT achieves the best scores.}
    \label{table:ablation_tasks}
    \vspace{-3mm}
\end{table}

\noindent{\textbf{Regarding NAPT Regularization:}}
We analyze the importance on the downstream performance of each regularization technique used during the NAPT phase. 
We report the performance in Table~\ref{table:ablation_reg}. 
We analyze the influence of: (i) Tuple Dropout, (ii) Biased weight decay, and (iii) Weight decay. 
We observe that our proposed approach is robust to hyperparameters, obtaining similar performance with 
various combinations of the $3$ regularization techniques. 
We attribute this to the way the NAPT dataset is split into train and validation: enforcing disjoint sets of aspect-terms. 
This allows us to detect when the model starts to overfit.\footnote{Preliminary experiments shows that regularization was needed, but the training and testing splits contained overlapping aspect terms and opinion terms.}
\begin{table*}[t]
    \centering
    \resizebox{1.0\textwidth}{!}{

\begin{tabular}{|c|c|c|c|}
\hline
\textbf{Task : Input} & \textbf{Gold} & \textbf{w/o NAPT} & \textbf{w/ NAPT} \\ \hline
\begin{tabular}[c]{@{}c@{}}\textbf{ASTE:} Given the text: Finally, the biggest problem has been tech\\support., what are the aspect terms and their sentiments?\end{tabular} & $<$tech support, \textit{negative}$>$ & $<$support, \textit{negative}$>$ & $<$tech support, \textit{negative}$>$ \\ \hline
\begin{tabular}[c]{@{}c@{}}\textbf{ASTE:} What are the aspect terms and their sentiments in the text: \\ Of course, for a student, weight is always an issue.?\end{tabular} & $<$weight, \textit{neutral}$>$ & $<$weight, \textit{neutral}$>$ & $<$weight, \textit{negative}$>$ \\ \hline
\begin{tabular}[c]{@{}c@{}}\textbf{AESC:} Given the text: the mouse buttons are hard to push., \\ what are the aspect term, opinion term, and sentiment triplets?\end{tabular} & $<$mouse buttons, hard, \textit{negative}$>$ & $<$ , , $>$ & $<$mouse buttons, hard, \textit{negative}$>$ \\ \hline
\begin{tabular}[c]{@{}c@{}}\textbf{AESC:} Given the text: The resolution is even higher then any other \\ laptop on the market., what are the aspect term, \\ opinion term and sentiment triplets?\end{tabular} & $<$resolution, higher, \textit{positive}$>$ & $<$resolution, higher, \textit{positive}$>$ & $<$laptop, higher, \textit{positive}$>$ \\ \hline
\end{tabular}%
    }
    \caption{Predictions made by an instruction tuned model with and without NAPT in low-shot scenarios.}
    \label{table:prediction_examples}
\end{table*}

\begin{table}[h]
    \centering
    \resizebox{0.5\textwidth}{!}{

\begin{tabular}{cccccc}
\hline
\multicolumn{3}{c}{\textbf{Ablation Config.}} & \multicolumn{3}{c}{\textbf{Dataset}} \\ \hline
\begin{tabular}[c]{@{}c@{}}Tuple \\ Dropout\end{tabular} & \begin{tabular}[c]{@{}c@{}}Weight \\ Decay\end{tabular} & \begin{tabular}[c]{@{}c@{}}Biased \\ Weight\end{tabular} & {\bf \textsc{Lap14}} & {\bf \textsc{Rest15}} & {\bf \textsc{Rest16}} \\ \hline
$\checkmark$ & $\checkmark$ & $\checkmark$ & $47.45$ & $47.32$ & $51.65$  \\
$\checkmark$ & $\checkmark$ & $\times$ & $47.57$ & $47.10$ & $51.39$  \\
$\checkmark$ & $\times$ & $\checkmark$ & $47.62$ & $47.26$ & $51.65$  \\
$\checkmark$ & $\times$ & $\times$ & $47.39$ & $47.17$ & $51.37$  \\
$\times$ & $\checkmark$ & $\checkmark$ & $47.55$ & $47.65$ & $51.80$  \\
$\times$ & $\checkmark$ & $\times$ & $46.43$ & $47.44$ & $51.49$  \\
$\times$ & $\times$ & $\checkmark$ & $46.78$ & $47.12$ & $51.11$  \\
$\times$ &$\times$ & $\times$ & $46.90$ & $47.27$ & $51.49$  \\ \hline
\end{tabular}
    }
    \caption{Ablation study over different regularization techniques in terms of macro F1 scores averaged across all tasks and $4$ $k$-shot settings.}
    \label{table:ablation_reg}
    \vspace{-3mm}
\end{table}

\subsection{Sentiment Prediction: Error Analysis}
\begin{figure}[t]
    \centering
    \includegraphics[width=0.49\textwidth]{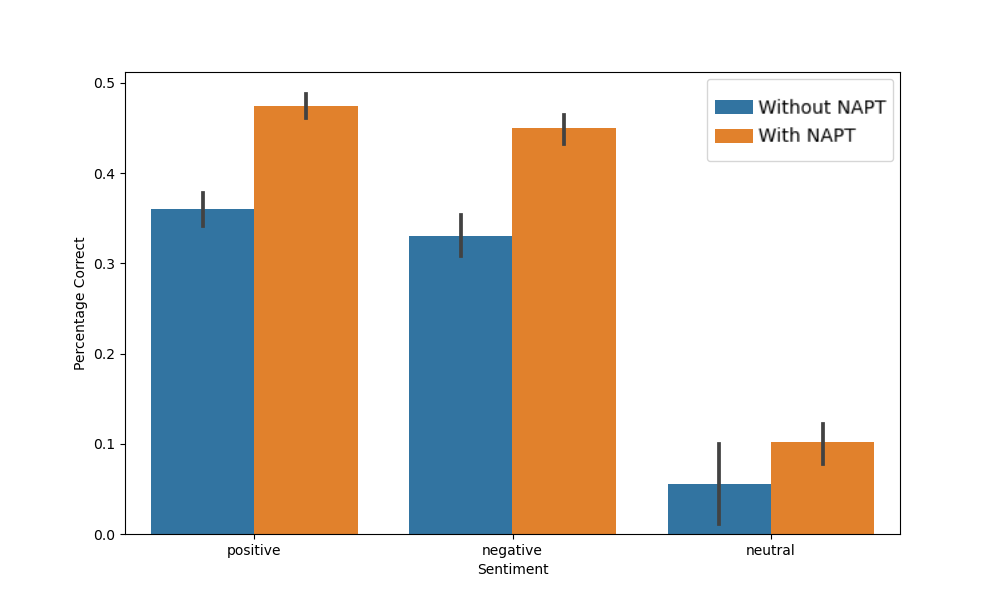}
    \caption{Comparison on the percentage of correct predictions over each sentiment class for an instruction tuned model with vs without the proposed NAPT on the \textsc{Lap14} dataset and $k=10$. With NAPT, it performs better on each sentiment class, even though \textit{neutral} class does not appear in the noisy dataset \tiny{(larger is better)}.} 
    \label{fig:sentiment_error}
\end{figure}

\textbf{Quantitative:} We first compare the percentage of correct predictions over each sentiment class, namely \textit{positive}, \textit{negative}, and \textit{neutral}. 
We compare instruction tuning with and without our proposed NAPT step.
We highlight the results in Figure~\ref{fig:sentiment_error}. 
We observe that our proposed method performs better for every sentiment class. 
Moreover, we note that our proposed method outperforms the baseline even for the \textit{neutral} sentiment class, a class which has not been seen during the NAPT phase.
This suggests that NAPT can help the model learn faster even unseen tasks.

\noindent{\textbf{Qualitative:}} We present examples of the predictions made by an instruction tuned model with and without our proposed NAPT in Table~\ref{table:prediction_examples}. 
We show $4$ predictions, $2$ for ASTE (first two rows) and $2$ for AESC (bottom two) on \texttt{LAP14}, in low-shot scenarios.
We observe that the baseline has difficulties extracting the full aspect term (first row), 
while our proposed method is able extract the complete triple. The metric used does not reward partial matching. In the second row, the baseline correctly generates the gold output, while our proposed method predicts a \textit{negative} sentiment.
In this case, the input can be considered ambiguous, as there is no explicit sentiment expressed in it.
Also, for more complex tasks, such as aspect sentiment triplet extraction (AESC), the baseline has difficulties generating a valid prediction, while our proposed method is able to generate the correct prediction (third row).
Lastly, we observe that although with NAPT we predict incorrectly (last row), it rather falls back to a term relevant to the domain (\textit{i.e.,} laptop). %, albeit incorrect.

\section{Conclusion}
\label{s:conclusion}

In this paper, we proposed to add an intermediate step in the \texttt{pretrain}$\rightarrow$\texttt{finetune} paradigm, called Noisy ABSA Pre-Training.
We motivate this newly introduced step with the hypothesis that exposing the model to tasks more aligned with the downstream task will improve
its performance, especially in low-data regimes such as in few-shot or complete zero-shot. We constructed a noisy dataset with a heuristic based pipeline approach consisting of three steps that utilize well-studied NLP resources and models. It serves as the training dataset for the noisy pre-training phase. We then evaluated with customer reviews from three datasets covering two domains, \textit{laptop} and \textit{restaurant}, and obtained large improvements in the zero/few-shot cases while achieving similar performance under finetuning on full dataset. We also discussed caveats around introducing catastrophic forgetting of general purpose pre-trained language models through such noisy pre-training, and introduced a few regularization techniques to help alleviate it.
\section*{Limitations}
\label{sec:limitations}
We believe our proposed noisy pre-training step should apply to other structured
prediction tasks, however we have not evaluated the approach on anything other than ABSA related tasks. Additionally, the noisy corpus construction process is heavily dependent on English based resources and pre-trained models. It might be non-trivial to extend the approach to other languages. Finally, we presented some extrinsic evaluation regarding the quality of the noisy corpus we create \textit{e.g.,} equivalence in terms of gold-annotated data size (Section \ref{sss:dataset_size_equivalent}). We leave any intrinsic evaluation of it by means of human supervision or otherwise for future work.

\bibliography{anthology,custom}
\bibliographystyle{acl_natbib}

\appendix
\section{Implementation details}
\label{appendix:implementation_details}
We use HuggingFace's implementation of transformers \cite{Wolf2020TransformersSN,Lhoest2021DatasetsAC}. We use similar parameters as \cite{Varia2022ABSAIT}. We run our experiments on NVIDIA Tesla V100 GPUs. 
\section{All Experiments}
\label{appendix:all_experiments}
For completeness, 
we include here all the models investigated over the 3 datasets, \textsc{Lap14}, \textsc{Rest15}, and \textsc{Rest16}, respectively.

\subsection{Full-Training}
\label{appendix_ss:full_training}
We report the results (\textit{test}) on Full Training in Tables~\ref{appendix_ss:full_training_lap14},~\ref{appendix_ss:full_training_rest15},~\ref{appendix_ss:full_training_rest16}.

\begin{table*}[h]
    \centering
    \resizebox{0.75\textwidth}{!}{
        \begin{tabular}{l|l|llll}
        \hline
            \multirow{2}{*}{Model}                                                                                                        & \multirow{2}{*}{NAPT} & \multicolumn{3}{c|}{Task (F1 $\uparrow$)}                          & \multirow{2}{*}{Average} \\
                                                                                                                                          &                      & AE            & AESC           & \multicolumn{1}{l|}{ASTE}           &         \\ \hline
            \begin{tabular}[c]{@{}l@{}}Text \\ \footnotesize{(t5-base)}\end{tabular}                                           & No                   & 76.13$\pm$1.06 & 66.57$\pm$1.01 & \multicolumn{1}{l|}{59.50$\pm$1.35} & 67.40$\pm$7.13 \\ \hline
            \multirow{2}{*}{\begin{tabular}[c]{@{}l@{}}IT \\ \footnotesize{(t5-base)}\end{tabular}}                       & No                   & 77.09$\pm$0.68 & 66.25$\pm$0.45 & \multicolumn{1}{l|}{60.47$\pm$1.36} & 67.94$\pm$7.18 \\
                                                                                                                                          & Yes                  & 76.96$\pm$1.17 & 66.08$\pm$0.80 & \multicolumn{1}{l|}{60.03$\pm$1.23} & 67.69$\pm$7.16 \\ \hline
            \multirow{2}{*}{\begin{tabular}[c]{@{}l@{}}IT-MTL \\ \footnotesize{(t5-base)}\end{tabular}}                 & No                   & 77.64$\pm$0.75 & 66.54$\pm$1.09 & \multicolumn{1}{l|}{60.17$\pm$1.19} & 68.11$\pm$7.53 \\
                                                                                                                                          & Yes                  & 77.67$\pm$1.04 & 66.66$\pm$0.69 & \multicolumn{1}{l|}{59.97$\pm$1.28} & 68.10$\pm$7.45 \\ \hline
            \multirow{2}{*}{\begin{tabular}[c]{@{}l@{}}IT \\ \footnotesize{(t5-large)}\end{tabular}}                      & No                   & 77.18$\pm$1.64 & 67.20$\pm$1.23 & \multicolumn{1}{l|}{60.24$\pm$0.61} & 68.21$\pm$7.28 \\
                                                                                                                                          & Yes                  & 76.79$\pm$1.05 & 66.66$\pm$1.16 & \multicolumn{1}{l|}{60.98$\pm$1.78} & 68.14$\pm$6.78 \\ \hline
            \multirow{2}{*}{\begin{tabular}[c]{@{}l@{}}IT-MTL \\ \footnotesize{(t5-large)}\end{tabular}}                & No                   & 77.89$\pm$0.53 & 66.44$\pm$1.06 & \multicolumn{1}{l|}{59.83$\pm$2.32} & 68.05$\pm$7.85 \\
                                                                                                                                          & Yes                  & 77.95$\pm$1.00 & 65.62$\pm$1.23 & \multicolumn{1}{l|}{59.34$\pm$1.42} & 67.64$\pm$7.95 \\ \hline
            IT                                                                                                            & No                   & 75.77$\pm$0.71 & 65.99$\pm$0.98 & \multicolumn{1}{l|}{59.28$\pm$0.64} & 67.01$\pm$7.05 \\
            \begin{tabular}[c]{@{}l@{}}\footnotesize{(continued pre-training)}\\ \footnotesize{(t5-base)}\end{tabular}             & Yes                  & 76.19$\pm$1.33 & 66.28$\pm$1.36 & \multicolumn{1}{l|}{59.38$\pm$1.25} & 67.28$\pm$7.09 \\ \hline 
            IT-MTL                                                                                                      & No                   & 76.37$\pm$0.82 & 65.85$\pm$1.03 & \multicolumn{1}{l|}{58.24$\pm$1.03} & 66.82$\pm$7.74 \\
            \begin{tabular}[c]{@{}l@{}}\footnotesize{(continued pre-training)}\\ \footnotesize{(t5-base)}\end{tabular}             & Yes                  & 76.68$\pm$0.88 & 65.95$\pm$1.06 & \multicolumn{1}{l|}{58.44$\pm$1.26} & 67.03$\pm$7.64 \\ \hline
        \end{tabular}
    }
    \caption{\footnotesize{Comparison of full dataset training performances on all $3$ ABSA tasks for \textsc{Lap14}.}}
    \label{appendix_ss:full_training_lap14}
    % \vspace{-5mm}
\end{table*}

\begin{table*}[h]
    \centering
    \resizebox{1.0\textwidth}{!}{
        \begin{tabular}{l|l|lllll|l}
        \hline
            \multirow{2}{*}{Model}                                                                                                        & \multirow{2}{*}{NAPT} & \multicolumn{5}{c|}{Task (F1 $\uparrow$)}                                                            & \multirow{2}{*}{Average} \\
                                                                                                                                          &                      & AE            & AESC           & TASD           & ASTE           & ASQP           &                          \\ \hline
            \begin{tabular}[c]{@{}l@{}}Text \\ \footnotesize{(t5-base)}\end{tabular}                                           & No                   & 72.76$\pm$0.96 & 66.43$\pm$1.45 & 60.05$\pm$0.67 & 51.74$\pm$0.84 & 46.66$\pm$0.67 & 59.53$\pm$9.72           \\ \hline
            \multirow{2}{*}{\begin{tabular}[c]{@{}l@{}}IT \\ \footnotesize{(t5-base)}\end{tabular}}                       & No                   & 73.54$\pm$1.20 & 67.09$\pm$0.53 & 59.78$\pm$0.91 & 52.78$\pm$0.81 & 46.79$\pm$0.59 & 59.99$\pm$9.82           \\
                                                                                                                                          & Yes                  & 72.89$\pm$1.31 & 65.98$\pm$1.29 & 59.30$\pm$0.77 & 52.62$\pm$1.13 & 46.49$\pm$0.71 & 59.45$\pm$9.48           \\ \hline
            \multirow{2}{*}{\begin{tabular}[c]{@{}l@{}}IT-MTL \\ \footnotesize{(t5-base)}\end{tabular}}                 & No                   & 73.85$\pm$1.14 & 67.46$\pm$0.80 & 59.88$\pm$1.02 & 53.17$\pm$0.67 & 47.17$\pm$1.03 & 60.30$\pm$9.81           \\
                                                                                                                                          & Yes                  & 74.55$\pm$1.26 & 67.53$\pm$1.37 & 59.29$\pm$1.67 & 53.57$\pm$1.42 & 47.30$\pm$1.21 & 60.45$\pm$9.86           \\ \hline
            \multirow{2}{*}{\begin{tabular}[c]{@{}l@{}}IT \\ \footnotesize{(t5-large)}\end{tabular}}                      & No                   & 74.24$\pm$0.74 & 69.83$\pm$1.10 & 62.82$\pm$0.69 & 55.96$\pm$0.41 & 49.61$\pm$0.55 & 62.49$\pm$9.16           \\
                                                                                                                                          & Yes                  & 74.68$\pm$0.72 & 69.94$\pm$1.18 & 62.82$\pm$0.94 & 54.72$\pm$1.53 & 49.48$\pm$1.04 & 62.33$\pm$9.47           \\ \hline
            \multirow{2}{*}{\begin{tabular}[c]{@{}l@{}}IT-MTL \\ \footnotesize{(t5-large)}\end{tabular}}                & No                   & 75.79$\pm$0.69 & 70.18$\pm$1.31 & 62.84$\pm$1.37 & 54.16$\pm$0.95 & 48.86$\pm$1.13 & 62.37$\pm$10.17          \\
                                                                                                                                          & Yes                  & 74.80$\pm$0.94 & 68.26$\pm$0.96 & 61.11$\pm$1.10 & 53.69$\pm$1.40 & 48.41$\pm$1.26 & 61.25$\pm$9.70           \\ \hline
            IT                                                                                                            & No                   & 73.05$\pm$1.05 & 67.17$\pm$1.16 & 59.09$\pm$0.91 & 51.89$\pm$1.09 & 46.51$\pm$0.36 & 59.54$\pm$9.92           \\
            \begin{tabular}[c]{@{}l@{}}\footnotesize{(continued pre-training)}\\ \footnotesize{(t5-base)}\end{tabular}             & Yes                  & 72.82$\pm$1.11 & 67.44$\pm$0.99 & 60.42$\pm$0.95 & 53.07$\pm$0.88 & 47.56$\pm$1.50 & 60.26$\pm$9.31           \\ \hline
            IT-MTL                                                                                                      & No                   & 74.14$\pm$0.47 & 68.06$\pm$0.49 & 60.97$\pm$0.59 & 53.42$\pm$1.27 & 47.49$\pm$0.90 & 60.82$\pm$9.84           \\
            \begin{tabular}[c]{@{}l@{}}\footnotesize{(continued pre-training)}\\ \footnotesize{(t5-base)}\end{tabular}             & Yes                  & 74.66$\pm$1.06 & 68.59$\pm$0.78 & 61.14$\pm$0.88 & 53.42$\pm$0.75 & 48.41$\pm$0.55 & 61.24$\pm$9.69           \\ \hline
        \end{tabular}
    }
    \caption{\footnotesize{Comparison of full dataset training performances on all $5$ ABSA tasks for \textsc{Rest15}.}}
    \label{appendix_ss:full_training_rest15}
\end{table*}

\begin{table*}[h]
    \centering
    \resizebox{1.0\textwidth}{!}{
        \begin{tabular}{l|l|lllll|l}
        \hline
            \multirow{2}{*}{Model}                                                                                                        & \multirow{2}{*}{NAPT} & \multicolumn{5}{c|}{Task (F1 $\uparrow$)}                                                            & \multirow{2}{*}{Average} \\
                                                                                                                                          &                      & AE            & AESC           & TASD           & ASTE           & ASQP           &                          \\ \hline
            \begin{tabular}[c]{@{}l@{}}Text \\ \footnotesize{(t5-base)}\end{tabular}                                           & No                   & 78.40$\pm$1.14 & 73.64$\pm$1.30 & 67.05$\pm$0.96 & 62.95$\pm$0.61 & 57.77$\pm$1.13 & 67.96$\pm$7.58           \\ \hline
            \multirow{2}{*}{\begin{tabular}[c]{@{}l@{}}IT \\ \footnotesize{(t5-base)}\end{tabular}}                       & No                   & 79.74$\pm$0.98 & 74.24$\pm$0.54 & 68.04$\pm$0.86 & 63.77$\pm$0.82 & 58.41$\pm$0.73 & 68.84$\pm$7.72           \\
                                                                                                                                          & Yes                  & 78.69$\pm$1.30 & 72.90$\pm$0.98 & 67.40$\pm$1.20 & 61.96$\pm$0.94 & 57.57$\pm$1.25 & 67.70$\pm$7.66           \\ \hline
            \multirow{2}{*}{\begin{tabular}[c]{@{}l@{}}IT-MTL \\ \footnotesize{(t5-base)}\end{tabular}}                 & No                   & 79.90$\pm$0.62 & 74.51$\pm$0.91 & 67.59$\pm$0.75 & 62.69$\pm$0.69 & 57.72$\pm$0.76 & 68.48$\pm$8.15           \\
                                                                                                                                          & Yes                  & 78.53$\pm$0.75 & 73.31$\pm$0.87 & 66.72$\pm$0.98 & 61.67$\pm$0.65 & 56.78$\pm$0.65 & 67.40$\pm$7.90           \\ \hline
            \multirow{2}{*}{\begin{tabular}[c]{@{}l@{}}IT \\ \footnotesize{(t5-large)}\end{tabular}}                      & No                   & 79.66$\pm$0.98 & 76.90$\pm$0.93 & 70.24$\pm$1.13 & 65.15$\pm$0.20 & 60.13$\pm$1.06 & 70.42$\pm$7.42           \\
                                                                                                                                          & Yes                  & 78.87$\pm$1.11 & 75.25$\pm$0.80 & 70.40$\pm$0.81 & 64.61$\pm$1.11 & 59.76$\pm$0.86 & 69.78$\pm$7.06           \\ \hline
            \multirow{2}{*}{\begin{tabular}[c]{@{}l@{}}IT-MTL \\ \footnotesize{(t5-large)}\end{tabular}}                & No                   & 79.67$\pm$0.50 & 75.01$\pm$0.95 & 69.12$\pm$1.04 & 62.84$\pm$0.98 & 58.79$\pm$0.99 & 69.09$\pm$7.85           \\
                                                                                                                                          & Yes                  & 79.33$\pm$0.78 & 74.66$\pm$0.72 & 67.11$\pm$1.66 & 62.43$\pm$0.99 & 57.17$\pm$1.17 & 68.14$\pm$8.18           \\ \hline
            IT                                                                                                            & No                   & 79.22$\pm$0.59 & 74.05$\pm$0.70 & 67.58$\pm$1.61 & 62.69$\pm$1.58 & 57.73$\pm$0.82 & 68.25$\pm$7.92           \\
            \begin{tabular}[c]{@{}l@{}}\footnotesize{(continued pre-training)}\\ \footnotesize{(t5-base)}\end{tabular}             & Yes                  & 79.06$\pm$0.92 & 74.38$\pm$1.30 & 68.40$\pm$1.21 & 62.33$\pm$1.25 & 58.24$\pm$0.83 & 68.48$\pm$7.74           \\ \hline
            IT-MTL                                                                                                      & No                   & 79.25$\pm$0.58 & 74.13$\pm$0.56 & 67.72$\pm$0.80 & 62.38$\pm$0.69 & 58.04$\pm$0.87 & 68.30$\pm$7.86           \\
            \begin{tabular}[c]{@{}l@{}}\footnotesize{(continued pre-training)}\\ \footnotesize{(t5-base)}\end{tabular}             & Yes                  & 78.72$\pm$0.73 & 73.88$\pm$0.95 & 67.16$\pm$1.00 & 62.00$\pm$1.15 & 56.61$\pm$1.01 & 67.68$\pm$8.05           \\ \hline
        \end{tabular}
    }
    \caption{\footnotesize{Comparison of full dataset training performances on all $5$ ABSA tasks for \textsc{Rest16}.}}
    \label{appendix_ss:full_training_rest16}
\end{table*}

\subsection{K-Shot Learning}
\label{appendix_ss:kshot_learning}
We report the results (\textit{test}) on K-Shot Learning in Tables~\ref{appendix_ss:kshot_lap14},~\ref{appendix_ss:kshot_rest15},~\ref{appendix_ss:kshot_rest16}.

\begin{table*}[h]
    \centering
    \resizebox{0.7\textwidth}{!}{
        \begin{tabular}{c|l|l|lll|l}
        \hline
            \multicolumn{1}{l|}{\multirow{2}{*}{K}} & \multirow{2}{*}{Model}                                                                                                        & \multirow{2}{*}{NAPT} & \multicolumn{3}{c|}{Task (F1 $\uparrow$)}              & \multirow{2}{*}{Average} \\
            \multicolumn{1}{l|}{}                   &                                                                                                                               &                      & AE        & AESC       & ASTE       &       \\ \hline
            \multirow{13}{*}{5}                     & \begin{tabular}[c]{@{}l@{}}Text \\ \footnotesize{(t5-base)}\end{tabular}                                           & No                   & 37.45$\pm$2.94	& 22.91$\pm$1.65	& 12.06$\pm$1.83	& 24.14$\pm$10.96           \\ \cline{2-7} 
                                                    & \multirow{2}{*}{\begin{tabular}[c]{@{}l@{}}IT \\ \footnotesize{(t5-base)}\end{tabular}}                       & No                   & 44.59$\pm$1.15	& 26.81$\pm$2.35	& 13.04$\pm$0.91	& 28.14$\pm$13.45           \\
                                                    &                                                                                                                               & Yes                  & 47.46$\pm$2.76	& 38.85$\pm$2.11	& 28.88$\pm$1.58	& 38.40$\pm$7.98          \\ \cline{2-7} 
                                                    & \multirow{2}{*}{\begin{tabular}[c]{@{}l@{}}IT-MTL \\ \footnotesize{(t5-base)}\end{tabular}}                 & No                   & 36.63$\pm$3.03	& 25.31$\pm$2.78	& 15.96$\pm$2.11	& 25.97$\pm$9.09          \\
                                                    &                                                                                                                               & Yes                  & 47.02$\pm$2.60	& 36.49$\pm$1.97	& 27.53$\pm$1.97	& 37.02$\pm$8.34          \\ \cline{2-7} 
                                                    & \multirow{2}{*}{\begin{tabular}[c]{@{}l@{}}IT \\ \footnotesize{(t5-large)}\end{tabular}}                      & No                   & 43.01$\pm$2.09	& 26.73$\pm$2.86	& 16.14$\pm$2.19	& 28.63$\pm$11.66           \\
                                                    &                                                                                                                               & Yes                  & 46.92$\pm$2.71	& 37.52$\pm$2.44	& 25.81$\pm$2.62	& 36.75$\pm$9.13          \\ \cline{2-7} 
                                                    & \multirow{2}{*}{\begin{tabular}[c]{@{}l@{}}IT-MTL \\ \footnotesize{(t5-large)}\end{tabular}}                & No                   & 40.88$\pm$3.65	& 27.47$\pm$2.72	& 17.37$\pm$2.51	& 28.57$\pm$10.35           \\
                                                    &                                                                                                                               & Yes                  & 45.30$\pm$3.29	& 32.47$\pm$5.05	& 23.54$\pm$5.34	& 33.77$\pm$10.13           \\ \cline{2-7} 
                                                    & IT                                                                                                            & No                   & 36.59$\pm$0.91	& 22.82$\pm$1.20	& 12.38$\pm$0.88	& 23.93$\pm$10.31           \\
                                                    & \begin{tabular}[c]{@{}l@{}}\footnotesize{(continued pre-training)}\\ \footnotesize{(t5-base)}\end{tabular}             & Yes                  & 45.83$\pm$1.80	& 38.85$\pm$1.31	& 28.15$\pm$1.84	& 37.61$\pm$7.53          \\ \cline{2-7} 
                                                    & IT-MTL                                                                                                      & No                   & 26.25$\pm$2.32	& 22.40$\pm$1.26	& 13.62$\pm$1.98	& 20.76$\pm$5.75          \\
                                                    & \begin{tabular}[c]{@{}l@{}}\footnotesize{(continued pre-training)}\\ \footnotesize{(t5-base)}\end{tabular}             & Yes                  & 45.28$\pm$1.27	& 36.61$\pm$1.46	& 27.33$\pm$2.02	& 36.41$\pm$7.58          \\ \hline
            \multirow{13}{*}{10}                    & \begin{tabular}[c]{@{}l@{}}Text \\ \footnotesize{(t5-base)}\end{tabular}                                           & No                   & 46.85$\pm$2.12	& 33.67$\pm$1.71	& 18.95$\pm$2.91	& 33.16$\pm$11.99           \\ \cline{2-7} 
                                                    & \multirow{2}{*}{\begin{tabular}[c]{@{}l@{}}IT \\ \footnotesize{(t5-base)}\end{tabular}}                       & No                   & 52.12$\pm$2.42	& 37.49$\pm$1.91	& 25.22$\pm$0.83	& 38.28$\pm$11.51           \\
                                                    &                                                                                                                               & Yes                  & 55.98$\pm$2.16	& 45.02$\pm$1.64	& 36.62$\pm$2.61	& 45.87$\pm$8.29          \\ \cline{2-7} 
                                                    & \multirow{2}{*}{\begin{tabular}[c]{@{}l@{}}IT-MTL \\ \footnotesize{(t5-base)}\end{tabular}}                 & No                   & 48.71$\pm$1.89	& 39.13$\pm$2.29	& 28.00$\pm$2.59	& 38.61$\pm$9.01          \\
                                                    &                                                                                                                               & Yes                  & 55.81$\pm$2.14	& 44.49$\pm$1.50	& 35.15$\pm$1.71	& 45.15$\pm$8.72          \\ \cline{2-7} 
                                                    & \multirow{2}{*}{\begin{tabular}[c]{@{}l@{}}IT \\ \footnotesize{(t5-large)}\end{tabular}}                      & No                   & 49.44$\pm$9.70	& 36.64$\pm$3.64	& 25.10$\pm$1.46	& 37.06$\pm$11.71           \\
                                                    &                                                                                                                               & Yes                  & 53.13$\pm$4.59	& 43.35$\pm$2.91	& 34.94$\pm$1.49	& 43.81$\pm$8.19          \\ \cline{2-7} 
                                                    & \multirow{2}{*}{\begin{tabular}[c]{@{}l@{}}IT-MTL \\ \footnotesize{(t5-large)}\end{tabular}}                & No                   & 49.23$\pm$4.91	& 36.13$\pm$2.07	& 27.16$\pm$3.74	& 37.51$\pm$10.01           \\
                                                    &                                                                                                                               & Yes                  & 51.99$\pm$3.47	& 41.45$\pm$2.28	& 31.05$\pm$4.58	& 41.50$\pm$9.35          \\ \cline{2-7} 
                                                    & IT                                                                                                            & No                   & 41.61$\pm$6.49	& 33.89$\pm$1.69	& 21.36$\pm$2.57	& 32.29$\pm$9.45          \\
                                                    & \begin{tabular}[c]{@{}l@{}}\footnotesize{(continued pre-training)}\\ \footnotesize{(t5-base)}\end{tabular}             & Yes                  & 55.69$\pm$2.27	& 45.77$\pm$1.55	& 34.51$\pm$1.20	& 45.32$\pm$8.91          \\ \cline{2-7} 
                                                    & IT-MTL                                                                                                      & No                   & 41.65$\pm$1.78	& 34.44$\pm$2.71	& 24.55$\pm$1.50	& 33.55$\pm$7.50          \\
                                                    & \begin{tabular}[c]{@{}l@{}}\footnotesize{(continued pre-training)}\\ \footnotesize{(t5-base)}\end{tabular}             & Yes                  & 56.16$\pm$2.60	& 46.17$\pm$1.79	& 35.25$\pm$1.06	& 45.86$\pm$8.84          \\ \hline
            \multirow{13}{*}{20}                    & \begin{tabular}[c]{@{}l@{}}Text \\ \footnotesize{(t5-base)}\end{tabular}                                           & No                   & 56.56$\pm$1.15	& 42.64$\pm$0.99	& 29.18$\pm$2.23	& 42.79$\pm$11.66           \\ \cline{2-7} 
                                                    & \multirow{2}{*}{\begin{tabular}[c]{@{}l@{}}IT \\ \footnotesize{(t5-base)}\end{tabular}}                       & No                   & 59.08$\pm$1.97	& 44.82$\pm$1.24	& 33.24$\pm$1.53	& 45.71$\pm$11.04           \\
                                                    &                                                                                                                               & Yes                  & 61.67$\pm$1.81	& 48.88$\pm$1.10	& 41.20$\pm$2.01	& 50.58$\pm$8.70          \\ \cline{2-7} 
                                                    & \multirow{2}{*}{\begin{tabular}[c]{@{}l@{}}IT-MTL \\ \footnotesize{(t5-base)}\end{tabular}}                 & No                   & 57.98$\pm$3.72	& 47.14$\pm$2.42	& 34.55$\pm$1.85	& 46.56$\pm$10.24           \\
                                                    &                                                                                                                               & Yes                  & 61.05$\pm$1.62	& 48.94$\pm$1.68	& 38.17$\pm$1.96	& 49.38$\pm$9.60          \\ \cline{2-7} 
                                                    & \multirow{2}{*}{\begin{tabular}[c]{@{}l@{}}IT \\ \footnotesize{(t5-large)}\end{tabular}}                      & No                   & 59.30$\pm$2.38	& 46.88$\pm$2.92	& 34.44$\pm$2.61	& 46.88$\pm$10.79           \\
                                                    &                                                                                                                               & Yes                  & 61.43$\pm$1.44	& 49.00$\pm$3.37	& 38.52$\pm$1.84	& 49.65$\pm$9.79          \\ \cline{2-7} 
                                                    & \multirow{2}{*}{\begin{tabular}[c]{@{}l@{}}IT-MTL \\ \footnotesize{(t5-large)}\end{tabular}}                & No                   & 61.02$\pm$2.89	& 46.78$\pm$4.32	& 36.00$\pm$1.17	& 47.93$\pm$10.99           \\
                                                    &                                                                                                                               & Yes                  & 61.16$\pm$1.97	& 49.68$\pm$2.13	& 38.10$\pm$2.41	& 49.65$\pm$9.80          \\ \cline{2-7} 
                                                    & IT                                                                                                            & No                   & 53.92$\pm$1.64	& 43.56$\pm$1.02	& 28.45$\pm$1.62	& 41.98$\pm$10.91           \\
                                                    & \begin{tabular}[c]{@{}l@{}}\footnotesize{(continued pre-training)}\\ \footnotesize{(t5-base)}\end{tabular}             & Yes                  & 60.06$\pm$2.47	& 49.73$\pm$1.48	& 40.19$\pm$1.64	& 49.99$\pm$8.42          \\ \cline{2-7} 
                                                    & IT-MTL                                                                                                      & No                   & 55.64$\pm$2.04	& 45.44$\pm$1.97	& 32.12$\pm$1.28	& 44.40$\pm$10.11           \\
                                                    & \begin{tabular}[c]{@{}l@{}}\footnotesize{(continued pre-training)}\\ \footnotesize{(t5-base)}\end{tabular}             & Yes                  & 60.93$\pm$1.36	& 49.85$\pm$1.65	& 37.96$\pm$1.78	& 49.58$\pm$9.61          \\ \hline
            \multirow{13}{*}{50}                    & \begin{tabular}[c]{@{}l@{}}Text \\ \footnotesize{(t5-base)}\end{tabular}                                           & No                   & 65.31$\pm$1.86	& 54.35$\pm$1.15	& 40.84$\pm$2.53	& 53.50$\pm$10.51           \\ \cline{2-7} 
                                                    & \multirow{2}{*}{\begin{tabular}[c]{@{}l@{}}IT \\ \footnotesize{(t5-base)}\end{tabular}}                       & No                   & 68.95$\pm$1.22	& 54.92$\pm$1.07	& 44.67$\pm$2.12	& 56.18$\pm$10.40           \\
                                                    &                                                                                                                               & Yes                  & 68.14$\pm$1.12	& 54.67$\pm$1.82	& 46.56$\pm$1.38	& 56.46$\pm$9.11          \\ \cline{2-7} 
                                                    & \multirow{2}{*}{\begin{tabular}[c]{@{}l@{}}IT-MTL \\ \footnotesize{(t5-base)}\end{tabular}}                 & No                   & 67.54$\pm$1.62	& 55.86$\pm$1.90	& 45.10$\pm$2.69	& 56.16$\pm$9.69          \\
                                                    &                                                                                                                               & Yes                  & 68.23$\pm$1.34	& 54.79$\pm$1.68	& 45.85$\pm$1.11	& 56.29$\pm$9.40          \\ \cline{2-7} 
                                                    & \multirow{2}{*}{\begin{tabular}[c]{@{}l@{}}IT \\ \footnotesize{(t5-large)}\end{tabular}}                      & No                   & 68.27$\pm$3.17	& 56.37$\pm$1.48	& 45.26$\pm$1.55	& 56.64$\pm$9.94          \\
                                                    &                                                                                                                               & Yes                  & 68.36$\pm$1.15	& 57.99$\pm$2.05	& 47.23$\pm$2.36	& 57.86$\pm$8.97          \\ \cline{2-7} 
                                                    & \multirow{2}{*}{\begin{tabular}[c]{@{}l@{}}IT-MTL \\ \footnotesize{(t5-large)}\end{tabular}}                & No                   & 69.92$\pm$1.23	& 56.33$\pm$1.24	& 44.87$\pm$2.10	& 57.04$\pm$10.70           \\
                                                    &                                                                                                                               & Yes                  & 70.07$\pm$1.30	& 55.99$\pm$0.95	& 45.99$\pm$2.25	& 57.35$\pm$10.16           \\ \cline{2-7} 
                                                    & IT                                                                                                            & No                   & 63.36$\pm$1.05	& 48.97$\pm$0.84	& 37.31$\pm$1.78	& 49.88$\pm$11.09           \\
                                                    & \begin{tabular}[c]{@{}l@{}}\footnotesize{(continued pre-training)}\\ \footnotesize{(t5-base)}\end{tabular}             & Yes                  & 68.78$\pm$1.42	& 55.20$\pm$1.08	& 45.50$\pm$1.44	& 56.49$\pm$9.74          \\ \cline{2-7} 
                                                    & IT-MTL                                                                                                      & No                   & 63.72$\pm$0.64	& 53.02$\pm$1.08	& 40.83$\pm$1.10	& 52.53$\pm$9.72          \\
                                                    & \begin{tabular}[c]{@{}l@{}}\footnotesize{(continued pre-training)}\\ \footnotesize{(t5-base)}\end{tabular}             & Yes                  & 69.19$\pm$1.31	& 55.73$\pm$1.11	& 45.44$\pm$1.56	& 56.79$\pm$9.92          \\ \hline
        \end{tabular}
    }
    \caption{\footnotesize{Comparison of $k$-Shot performances on all $3$ ABSA tasks for \textsc{Lap14}.}}
    \label{appendix_ss:kshot_lap14}
\end{table*}

\begin{table*}[h]
    \centering
    \resizebox{0.84\textwidth}{!}{
        \begin{tabular}{c|l|l|lllll|l}
        \hline
            \multicolumn{1}{l|}{\multirow{2}{*}{K}} & \multirow{2}{*}{Model}                                                                          & \multirow{2}{*}{NAPT} & \multicolumn{5}{c|}{Task (F1 $\uparrow$)}                                        & \multirow{2}{*}{Average} \\
            \multicolumn{1}{l|}{}                   &                                                                                                 &                      & ATE        & AESC       & TASD       & ASTE       & ASQP       &                          \\ \hline
            \multirow{13}{*}{5}                     & \begin{tabular}[c]{@{}l@{}}Text \\ (t5-base)\end{tabular}                            & No                   & 44.55$\pm$2.55	& 39.44$\pm$2.64	& 24.62$\pm$1.56	& 20.11$\pm$1.05	& 12.88$\pm$0.91	& 28.32$\pm$12.26             \\ \cline{2-9} 
                                                    & \multirow{2}{*}{\begin{tabular}[c]{@{}l@{}}IT \\ (t5-base)\end{tabular}}        & No                   & 49.33$\pm$0.66	& 42.48$\pm$1.84	& 24.75$\pm$0.65	& 24.44$\pm$1.09	& 15.52$\pm$1.47	& 31.31$\pm$12.87             \\
                                                    &                                                                                                 & Yes                  & 50.05$\pm$2.91	& 43.95$\pm$1.79	& 30.46$\pm$1.87	& 31.59$\pm$1.35	& 21.72$\pm$0.90	& 35.56$\pm$10.37             \\ \cline{2-9} 
                                                    & \multirow{2}{*}{\begin{tabular}[c]{@{}l@{}}IT-MTL \\ (t5-base)\end{tabular}}  & No                   & 48.14$\pm$2.79	& 41.42$\pm$3.28	& 24.79$\pm$2.33	& 24.49$\pm$1.85	& 15.28$\pm$1.64	& 30.82$\pm$12.53             \\
                                                    &                                                                                                 & Yes                  & 51.11$\pm$1.81	& 43.51$\pm$1.55	& 27.12$\pm$1.97	& 30.35$\pm$1.48	& 18.98$\pm$1.39	& 34.21$\pm$11.76             \\ \cline{2-9} 
                                                    & \multirow{2}{*}{\begin{tabular}[c]{@{}l@{}}IT \\ (t5-large)\end{tabular}}       & No                   & 46.40$\pm$1.56	& 41.24$\pm$0.86	& 24.73$\pm$1.99	& 22.72$\pm$1.95	& 16.04$\pm$3.00	& 30.23$\pm$11.96             \\
                                                    &                                                                                                 & Yes                  & 47.87$\pm$4.76	& 43.01$\pm$2.77	& 28.42$\pm$7.70	& 30.49$\pm$1.43	& 20.85$\pm$1.79	& 34.13$\pm$10.84             \\ \cline{2-9} 
                                                    & \multirow{2}{*}{\begin{tabular}[c]{@{}l@{}}IT-MTL \\ (t5-large)\end{tabular}} & No                   & 44.54$\pm$2.84	& 36.25$\pm$1.78	& 19.08$\pm$3.03	& 18.92$\pm$3.92	& 10.57$\pm$2.01	& 25.87$\pm$13.05             \\
                                                    &                                                                                                 & Yes                  & 48.47$\pm$1.98	& 40.38$\pm$2.76	& 23.79$\pm$3.88	& 26.97$\pm$3.56	& 16.25$\pm$3.37	& 31.17$\pm$12.16             \\ \cline{2-9} 
                                                    & IT                                                                              & No                   & 46.06$\pm$2.36	& 39.34$\pm$3.07	& 24.67$\pm$1.17	& 22.70$\pm$0.85	& 14.47$\pm$1.62	& 29.45$\pm$11.92             \\
                                                    & \begin{tabular}[c]{@{}l@{}}(continued pre-training) \\ (t5-base)\end{tabular}            & Yes                  & 50.40$\pm$1.76	& 44.06$\pm$1.59	& 29.32$\pm$2.16	& 31.31$\pm$2.31	& 22.20$\pm$2.32	& 35.46$\pm$10.53             \\ \cline{2-9} 
                                                    & IT-MTL                                                                        & No                   & 47.78$\pm$2.49	& 39.59$\pm$1.24	& 24.33$\pm$1.43	& 22.93$\pm$0.56	& 14.55$\pm$1.32	& 29.84$\pm$12.40             \\
                                                    & \begin{tabular}[c]{@{}l@{}}(continued pre-training)\\ (t5-base)\end{tabular}             & Yes                  & 50.87$\pm$2.76	& 44.15$\pm$2.18	& 29.30$\pm$2.79	& 31.60$\pm$2.05	& 20.98$\pm$2.28	& 35.38$\pm$11.06             \\ \hline
            \multirow{13}{*}{10}                    & \begin{tabular}[c]{@{}l@{}}Text \\ (t5-base)\end{tabular}                            & No                   & 54.71$\pm$0.91	& 49.28$\pm$0.46	& 36.26$\pm$1.62	& 31.99$\pm$0.80	& 24.42$\pm$0.68	& 39.33$\pm$11.41             \\ \cline{2-9} 
                                                    & \multirow{2}{*}{\begin{tabular}[c]{@{}l@{}}IT \\ (t5-base)\end{tabular}}        & No                   & 56.62$\pm$1.59	& 51.03$\pm$1.93	& 37.64$\pm$1.50	& 33.25$\pm$1.54	& 25.76$\pm$1.08	& 40.86$\pm$11.71             \\
                                                    &                                                                                                 & Yes                  & 57.91$\pm$1.29	& 50.78$\pm$1.42	& 37.37$\pm$1.81	& 37.63$\pm$1.26	& 28.78$\pm$1.11	& 42.49$\pm$10.59             \\ \cline{2-9} 
                                                    & \multirow{2}{*}{\begin{tabular}[c]{@{}l@{}}IT-MTL \\ (t5-base)\end{tabular}}  & No                   & 58.10$\pm$0.72	& 48.27$\pm$0.98	& 37.26$\pm$0.29	& 33.75$\pm$0.74	& 26.48$\pm$1.01	& 40.77$\pm$11.41             \\
                                                    &                                                                                                 & Yes                  & 58.72$\pm$1.23	& 49.95$\pm$1.30	& 36.77$\pm$1.68	& 37.82$\pm$1.70	& 28.03$\pm$1.21	& 42.26$\pm$10.95             \\ \cline{2-9} 
                                                    & \multirow{2}{*}{\begin{tabular}[c]{@{}l@{}}IT \\ (t5-large)\end{tabular}}       & No                   & 54.58$\pm$1.99	& 48.32$\pm$1.27	& 35.31$\pm$1.90	& 34.55$\pm$0.86	& 25.43$\pm$1.79	& 39.64$\pm$10.76             \\
                                                    &                                                                                                 & Yes                  & 55.69$\pm$1.94	& 49.52$\pm$1.42	& 38.11$\pm$1.76	& 36.54$\pm$1.71	& 28.10$\pm$1.53	& 41.59$\pm$10.04             \\ \cline{2-9} 
                                                    & \multirow{2}{*}{\begin{tabular}[c]{@{}l@{}}IT-MTL \\ (t5-large)\end{tabular}} & No                   & 54.14$\pm$1.11	& 45.38$\pm$1.09	& 33.90$\pm$2.76	& 30.95$\pm$1.68	& 23.10$\pm$1.47	& 37.49$\pm$11.31             \\
                                                    &                                                                                                 & Yes                  & 55.00$\pm$3.53	& 46.91$\pm$3.01	& 35.09$\pm$2.65	& 32.82$\pm$2.97	& 24.79$\pm$2.71	& 38.92$\pm$11.19             \\ \cline{2-9} 
                                                    & IT                                                                              & No                   & 56.55$\pm$2.35	& 51.28$\pm$0.82	& 39.02$\pm$2.58	& 33.70$\pm$1.41	& 25.10$\pm$0.66	& 41.13$\pm$11.81             \\
                                                    & \begin{tabular}[c]{@{}l@{}}(continued pre-training) \\ (t5-base)\end{tabular}            & Yes                  & 57.96$\pm$1.36	& 51.42$\pm$1.41	& 39.33$\pm$1.29	& 37.81$\pm$1.68	& 29.57$\pm$1.32	& 43.22$\pm$10.31             \\ \cline{2-9} 
                                                    & IT-MTL                                                                        & No                   & 58.31$\pm$0.92	& 49.57$\pm$2.13	& 39.00$\pm$2.28	& 33.01$\pm$1.21	& 25.58$\pm$0.75	& 41.09$\pm$11.98             \\
                                                    & \begin{tabular}[c]{@{}l@{}}(continued pre-training)\\ (t5-base)\end{tabular}             & Yes                  & 57.88$\pm$1.58	& 50.34$\pm$1.87	& 38.56$\pm$1.47	& 37.83$\pm$1.22	& 28.73$\pm$1.32	& 42.67$\pm$10.42             \\ \hline
            \multirow{13}{*}{20}                    & \begin{tabular}[c]{@{}l@{}}Text \\ (t5-base)\end{tabular}                            & No                   & 58.91$\pm$1.69	& 53.77$\pm$0.90	& 42.37$\pm$1.55	& 37.27$\pm$1.85	& 30.45$\pm$0.83	& 44.55$\pm$10.76             \\ \cline{2-9} 
                                                    & \multirow{2}{*}{\begin{tabular}[c]{@{}l@{}}IT \\ (t5-base)\end{tabular}}        & No                   & 62.08$\pm$1.85	& 53.91$\pm$2.18	& 42.89$\pm$0.86	& 38.35$\pm$0.83	& 30.77$\pm$1.19	& 45.60$\pm$11.45             \\
                                                    &                                                                                                 & Yes                  & 61.84$\pm$1.18	& 53.80$\pm$1.19	& 44.13$\pm$1.19	& 41.93$\pm$1.13	& 34.23$\pm$1.30	& 47.19$\pm$9.76             \\ \cline{2-9} 
                                                    & \multirow{2}{*}{\begin{tabular}[c]{@{}l@{}}IT-MTL \\ (t5-base)\end{tabular}}  & No                   & 63.77$\pm$1.86	& 53.47$\pm$2.10	& 43.27$\pm$1.33	& 40.66$\pm$2.07	& 33.27$\pm$0.76	& 46.89$\pm$10.97             \\
                                                    &                                                                                                 & Yes                  & 63.77$\pm$1.15	& 55.48$\pm$1.55	& 44.24$\pm$1.18	& 42.77$\pm$1.16	& 34.71$\pm$0.91	& 48.19$\pm$10.36             \\ \cline{2-9} 
                                                    & \multirow{2}{*}{\begin{tabular}[c]{@{}l@{}}IT \\ (t5-large)\end{tabular}}       & No                   & 59.97$\pm$1.49	& 55.11$\pm$1.86	& 45.59$\pm$1.00	& 40.27$\pm$1.10	& 34.40$\pm$1.80	& 47.07$\pm$9.67             \\
                                                    &                                                                                                 & Yes                  & 62.13$\pm$1.32	& 55.85$\pm$1.68	& 46.35$\pm$2.68	& 41.79$\pm$0.71	& 35.69$\pm$1.19	& 48.36$\pm$9.75             \\ \cline{2-9} 
                                                    & \multirow{2}{*}{\begin{tabular}[c]{@{}l@{}}IT-MTL \\ (t5-large)\end{tabular}} & No                   & 62.26$\pm$1.55	& 54.59$\pm$2.62	& 45.04$\pm$1.44	& 40.39$\pm$2.01	& 34.23$\pm$1.12	& 47.30$\pm$10.35             \\
                                                    &                                                                                                 & Yes                  & 63.19$\pm$1.70	& 55.67$\pm$2.23	& 44.23$\pm$1.40	& 41.77$\pm$1.48	& 34.43$\pm$1.25	& 47.86$\pm$10.49             \\ \cline{2-9} 
                                                    & IT                                                                              & No                   & 62.30$\pm$1.44	& 55.82$\pm$1.49	& 45.16$\pm$1.25	& 38.23$\pm$1.54	& 31.58$\pm$0.96	& 46.62$\pm$11.52             \\
                                                    & \begin{tabular}[c]{@{}l@{}}(continued pre-training) \\ (t5-base)\end{tabular}            & Yes                  & 62.85$\pm$1.38	& 56.12$\pm$0.90	& 45.51$\pm$1.57	& 42.07$\pm$1.53	& 34.48$\pm$1.13	& 48.21$\pm$10.25             \\ \cline{2-9} 
                                                    & IT-MTL                                                                        & No                   & 63.42$\pm$0.89	& 55.09$\pm$0.49	& 46.43$\pm$1.13	& 40.40$\pm$1.45	& 32.85$\pm$0.67	& 47.64$\pm$11.00             \\
                                                    & \begin{tabular}[c]{@{}l@{}}(continued pre-training)\\ (t5-base)\end{tabular}             & Yes                  & 63.91$\pm$1.21	& 56.14$\pm$1.47	& 46.40$\pm$1.18	& 42.80$\pm$1.34	& 36.15$\pm$0.92	& 49.08$\pm$9.97             \\ \hline
            \multirow{13}{*}{50}                    & \begin{tabular}[c]{@{}l@{}}Text \\ (t5-base)\end{tabular}                            & No                   & 62.55$\pm$1.74	& 57.12$\pm$1.31	& 48.50$\pm$0.97	& 43.09$\pm$0.91	& 35.51$\pm$0.82	& 49.35$\pm$9.91             \\ \cline{2-9} 
                                                    & \multirow{2}{*}{\begin{tabular}[c]{@{}l@{}}IT \\ (t5-base)\end{tabular}}        & No                   & 64.74$\pm$1.15	& 59.35$\pm$0.91	& 50.40$\pm$0.65	& 43.79$\pm$1.12	& 37.51$\pm$0.72	& 51.16$\pm$10.17             \\
                                                    &                                                                                                 & Yes                  & 65.17$\pm$0.76	& 58.96$\pm$0.92	& 49.72$\pm$1.24	& 44.74$\pm$1.34	& 39.10$\pm$1.16	& 51.54$\pm$9.56             \\ \cline{2-9} 
                                                    & \multirow{2}{*}{\begin{tabular}[c]{@{}l@{}}IT-MTL \\ (t5-base)\end{tabular}}  & No                   & 67.51$\pm$0.89	& 58.98$\pm$1.52	& 50.45$\pm$1.49	& 45.27$\pm$0.76	& 37.69$\pm$1.04	& 51.98$\pm$10.68             \\
                                                    &                                                                                                 & Yes                  & 67.55$\pm$1.18	& 60.19$\pm$1.23	& 50.51$\pm$1.09	& 46.76$\pm$0.93	& 39.94$\pm$0.86	& 52.99$\pm$9.91             \\ \cline{2-9} 
                                                    & \multirow{2}{*}{\begin{tabular}[c]{@{}l@{}}IT \\ (t5-large)\end{tabular}}       & No                   & 64.75$\pm$0.94	& 59.33$\pm$0.47	& 52.19$\pm$0.93	& 45.59$\pm$0.75	& 40.66$\pm$1.12	& 52.50$\pm$8.99             \\
                                                    &                                                                                                 & Yes                  & 66.82$\pm$1.16	& 61.21$\pm$1.40	& 52.53$\pm$1.76	& 47.19$\pm$1.30	& 42.27$\pm$1.41	& 54.00$\pm$9.16             \\ \cline{2-9} 
                                                    & \multirow{2}{*}{\begin{tabular}[c]{@{}l@{}}IT-MTL \\ (t5-large)\end{tabular}} & No                   & 67.84$\pm$1.16	& 60.77$\pm$1.23	& 51.70$\pm$0.97	& 46.76$\pm$1.45	& 39.92$\pm$1.00	& 53.40$\pm$10.18             \\
                                                    &                                                                                                 & Yes                  & 68.15$\pm$0.86	& 61.67$\pm$0.94	& 52.02$\pm$1.47	& 47.33$\pm$1.30	& 41.24$\pm$1.18	& 54.08$\pm$9.86             \\ \cline{2-9} 
                                                    & IT                                                                              & No                   & 64.49$\pm$0.95	& 60.23$\pm$0.51	& 51.51$\pm$0.81	& 44.10$\pm$1.74	& 37.56$\pm$1.30	& 51.58$\pm$10.20             \\
                                                    & \begin{tabular}[c]{@{}l@{}}(continued pre-training) \\ (t5-base)\end{tabular}            & Yes                  & 65.37$\pm$1.20	& 59.64$\pm$1.12	& 51.08$\pm$0.65	& 45.49$\pm$1.14	& 39.37$\pm$0.90	& 52.19$\pm$9.49             \\ \cline{2-9} 
                                                    & IT-MTL                                                                        & No                   & 67.46$\pm$1.03	& 61.93$\pm$0.70	& 52.73$\pm$1.10	& 46.06$\pm$0.61	& 39.71$\pm$1.70	& 53.58$\pm$10.38             \\
                                                    & \begin{tabular}[c]{@{}l@{}}(continued pre-training)\\ (t5-base)\end{tabular}             & Yes                  & 67.37$\pm$0.96	& 60.54$\pm$1.29	& 51.57$\pm$1.25	& 46.96$\pm$1.12	& 40.39$\pm$1.03	& 53.37$\pm$9.72             \\ \hline
        \end{tabular}
    }
    \caption{\footnotesize{Comparison of $k$-Shot performances on all $5$ ABSA tasks for \textsc{Rest15}.}}
    \label{appendix_ss:kshot_rest15}
\end{table*}

\begin{table*}[h]
    \centering
    \resizebox{0.84\textwidth}{!}{
        \begin{tabular}{c|l|l|lllll|l}
        \hline
            \multicolumn{1}{l|}{\multirow{2}{*}{K}} & \multirow{2}{*}{Model}                                                                          & \multirow{2}{*}{NAPT} & \multicolumn{5}{c|}{Task (F1 $\uparrow$)}                                        & \multirow{2}{*}{Average} \\
            \multicolumn{1}{l|}{}                   &                                                                                                 &                      & AE        & AESC       & TASD       & ASTE       & ASQP       &                          \\ \hline
            \multirow{13}{*}{5}                     & \begin{tabular}[c]{@{}l@{}}Text \\ (t5-base)\end{tabular}                            & No                   & 52.67$\pm$0.69	& 47.87$\pm$1.34	& 31.57$\pm$1.74	& 29.58$\pm$1.96	& 19.76$\pm$1.44	& 36.29$\pm$12.52             \\ \cline{2-9} 
                                                    & \multirow{2}{*}{\begin{tabular}[c]{@{}l@{}}IT \\ (t5-base)\end{tabular}}        & No                   & 55.59$\pm$2.74	& 51.62$\pm$1.46	& 36.26$\pm$1.15	& 34.10$\pm$1.17	& 23.89$\pm$2.11	& 40.29$\pm$12.07             \\
                                                    &                                                                                                 & Yes                  & 61.54$\pm$1.35	& 55.32$\pm$2.05	& 39.13$\pm$2.11	& 40.18$\pm$1.60	& 28.64$\pm$1.82	& 44.96$\pm$12.09             \\ \cline{2-9} 
                                                    & \multirow{2}{*}{\begin{tabular}[c]{@{}l@{}}IT-MTL \\ (t5-base)\end{tabular}}  & No                   & 59.78$\pm$1.32	& 52.35$\pm$0.82	& 36.88$\pm$1.77	& 36.27$\pm$0.90	& 25.86$\pm$1.63	& 42.23$\pm$12.50             \\
                                                    &                                                                                                 & Yes                  & 64.25$\pm$1.60	& 55.22$\pm$1.35	& 38.97$\pm$2.19	& 40.95$\pm$1.36	& 29.58$\pm$1.76	& 45.79$\pm$12.54             \\ \cline{2-9} 
                                                    & \multirow{2}{*}{\begin{tabular}[c]{@{}l@{}}IT \\ (t5-large)\end{tabular}}       & No                   & 55.88$\pm$1.63	& 52.90$\pm$2.02	& 38.37$\pm$2.79	& 36.70$\pm$0.83	& 27.70$\pm$1.85	& 42.31$\pm$10.91             \\
                                                    &                                                                                                 & Yes                  & 62.01$\pm$1.48	& 55.91$\pm$2.68	& 37.09$\pm$7.90	& 41.14$\pm$1.78	& 32.13$\pm$2.04	& 45.66$\pm$12.13             \\ \cline{2-9} 
                                                    & \multirow{2}{*}{\begin{tabular}[c]{@{}l@{}}IT-MTL \\ (t5-large)\end{tabular}} & No                   & 56.81$\pm$2.44	& 48.65$\pm$1.32	& 32.64$\pm$2.56	& 32.47$\pm$1.80	& 23.36$\pm$1.16	& 38.79$\pm$12.52             \\
                                                    &                                                                                                 & Yes                  & 60.50$\pm$1.91	& 51.89$\pm$2.50	& 34.94$\pm$3.71	& 37.71$\pm$2.08	& 27.04$\pm$2.22	& 42.42$\pm$12.46             \\ \cline{2-9} 
                                                    & IT                                                                              & No                   & 55.87$\pm$2.42	& 50.92$\pm$3.05	& 36.57$\pm$1.38	& 31.41$\pm$1.94	& 20.39$\pm$2.38	& 39.03$\pm$13.37             \\
                                                    & \begin{tabular}[c]{@{}l@{}}(continued pre-training) \\ (t5-base)\end{tabular}            & Yes                  & 62.22$\pm$1.99	& 56.70$\pm$1.43	& 37.00$\pm$2.16	& 39.19$\pm$1.67	& 27.18$\pm$1.64	& 44.46$\pm$13.22             \\ \cline{2-9} 
                                                    & IT-MTL                                                                        & No                   & 55.93$\pm$1.71	& 48.95$\pm$2.26	& 34.71$\pm$2.20	& 32.02$\pm$0.88	& 22.79$\pm$1.46	& 38.88$\pm$12.31             \\
                                                    & \begin{tabular}[c]{@{}l@{}}(continued pre-training)\\ (t5-base)\end{tabular}             & Yes                  & 62.74$\pm$1.21	& 55.43$\pm$0.86	& 37.13$\pm$2.36	& 39.84$\pm$1.37	& 27.80$\pm$1.88	& 44.59$\pm$12.89             \\ \hline
            \multirow{13}{*}{10}                    & \begin{tabular}[c]{@{}l@{}}Text \\ (t5-base)\end{tabular}                            & No                   & 59.45$\pm$0.89	& 54.33$\pm$1.03	& 38.85$\pm$1.95	& 36.82$\pm$0.91	& 29.31$\pm$1.17	& 43.75$\pm$11.59             \\ \cline{2-9} 
                                                    & \multirow{2}{*}{\begin{tabular}[c]{@{}l@{}}IT \\ (t5-base)\end{tabular}}        & No                   & 62.14$\pm$1.14	& 57.02$\pm$2.17	& 40.34$\pm$2.22	& 40.37$\pm$0.74	& 29.90$\pm$0.94	& 45.95$\pm$12.20             \\
                                                    &                                                                                                 & Yes                  & 65.33$\pm$1.18	& 58.84$\pm$1.48	& 42.69$\pm$2.83	& 44.24$\pm$1.07	& 32.30$\pm$1.39	& 48.68$\pm$12.07             \\ \cline{2-9} 
                                                    & \multirow{2}{*}{\begin{tabular}[c]{@{}l@{}}IT-MTL \\ (t5-base)\end{tabular}}  & No                   & 64.03$\pm$1.81	& 56.51$\pm$0.97	& 41.53$\pm$1.12	& 39.66$\pm$1.50	& 31.27$\pm$1.37	& 46.60$\pm$12.24             \\
                                                    &                                                                                                 & Yes                  & 65.85$\pm$1.08	& 57.96$\pm$1.14	& 41.66$\pm$2.32	& 44.42$\pm$1.00	& 32.77$\pm$2.38	& 48.53$\pm$12.04             \\ \cline{2-9} 
                                                    & \multirow{2}{*}{\begin{tabular}[c]{@{}l@{}}IT \\ (t5-large)\end{tabular}}       & No                   & 59.01$\pm$1.07	& 51.11$\pm$3.59	& 42.76$\pm$1.89	& 39.66$\pm$1.81	& 31.75$\pm$2.54	& 44.86$\pm$9.84             \\
                                                    &                                                                                                 & Yes                  & 61.41$\pm$2.08	& 57.90$\pm$1.18	& 43.13$\pm$2.28	& 43.26$\pm$1.74	& 35.40$\pm$2.29	& 48.22$\pm$10.10             \\ \cline{2-9} 
                                                    & \multirow{2}{*}{\begin{tabular}[c]{@{}l@{}}IT-MTL \\ (t5-large)\end{tabular}} & No                   & 59.76$\pm$1.11	& 53.26$\pm$2.04	& 39.01$\pm$2.52	& 37.45$\pm$1.46	& 29.06$\pm$1.24	& 43.71$\pm$11.52             \\
                                                    &                                                                                                 & Yes                  & 61.85$\pm$1.89	& 54.15$\pm$2.23	& 39.64$\pm$2.10	& 39.74$\pm$2.18	& 31.13$\pm$2.01	& 45.30$\pm$11.39             \\ \cline{2-9} 
                                                    & IT                                                                              & No                   & 59.25$\pm$2.32	& 56.57$\pm$2.06	& 39.28$\pm$2.12	& 37.84$\pm$1.59	& 26.17$\pm$1.79	& 43.82$\pm$12.78             \\
                                                    & \begin{tabular}[c]{@{}l@{}}(continued pre-training) \\ (t5-base)\end{tabular}            & Yes                  & 63.34$\pm$2.30	& 59.95$\pm$1.25	& 42.75$\pm$2.43	& 44.85$\pm$2.03	& 32.25$\pm$1.23	& 48.63$\pm$11.73             \\ \cline{2-9} 
                                                    & IT-MTL                                                                        & No                   & 60.50$\pm$1.25	& 55.34$\pm$0.67	& 41.57$\pm$2.03	& 38.22$\pm$0.89	& 30.40$\pm$1.30	& 45.20$\pm$11.41             \\
                                                    & \begin{tabular}[c]{@{}l@{}}(continued pre-training)\\ (t5-base)\end{tabular}             & Yes                  & 65.10$\pm$1.28	& 57.91$\pm$1.32	& 43.31$\pm$1.75	& 43.55$\pm$1.59	& 34.27$\pm$1.53	& 48.83$\pm$11.28             \\ \hline
            \multirow{13}{*}{20}                    & \begin{tabular}[c]{@{}l@{}}Text \\ (t5-base)\end{tabular}                            & No                   & 63.34$\pm$1.24	& 57.56$\pm$1.21	& 44.90$\pm$1.99	& 42.11$\pm$1.82	& 35.20$\pm$0.58	& 48.62$\pm$10.62             \\ \cline{2-9} 
                                                    & \multirow{2}{*}{\begin{tabular}[c]{@{}l@{}}IT \\ (t5-base)\end{tabular}}        & No                   & 65.89$\pm$1.90	& 60.52$\pm$1.44	& 47.27$\pm$2.49	& 44.27$\pm$0.99	& 36.39$\pm$0.75	& 50.87$\pm$11.14             \\
                                                    &                                                                                                 & Yes                  & 66.73$\pm$1.49	& 60.78$\pm$1.11	& 50.49$\pm$1.21	& 47.75$\pm$1.07	& 40.14$\pm$1.28	& 53.18$\pm$9.61             \\ \cline{2-9} 
                                                    & \multirow{2}{*}{\begin{tabular}[c]{@{}l@{}}IT-MTL \\ (t5-base)\end{tabular}}  & No                   & 65.82$\pm$0.96	& 59.66$\pm$1.06	& 49.30$\pm$0.99	& 44.71$\pm$0.72	& 38.71$\pm$0.76	& 51.64$\pm$10.10             \\
                                                    &                                                                                                 & Yes                  & 67.97$\pm$0.97	& 60.81$\pm$1.05	& 49.82$\pm$1.09	& 47.94$\pm$1.11	& 40.25$\pm$1.24	& 53.36$\pm$9.95             \\ \cline{2-9} 
                                                    & \multirow{2}{*}{\begin{tabular}[c]{@{}l@{}}IT \\ (t5-large)\end{tabular}}       & No                   & 64.63$\pm$0.41	& 61.07$\pm$0.94	& 49.74$\pm$2.34	& 46.02$\pm$1.34	& 40.53$\pm$1.04	& 52.40$\pm$9.36             \\
                                                    &                                                                                                 & Yes                  & 65.24$\pm$1.28	& 60.14$\pm$2.72	& 51.82$\pm$1.85	& 48.44$\pm$0.97	& 41.14$\pm$1.09	& 53.35$\pm$8.76             \\ \cline{2-9} 
                                                    & \multirow{2}{*}{\begin{tabular}[c]{@{}l@{}}IT-MTL \\ (t5-large)\end{tabular}} & No                   & 66.26$\pm$2.38	& 59.48$\pm$2.01	& 48.37$\pm$2.96	& 44.70$\pm$3.16	& 37.42$\pm$3.04	& 51.25$\pm$10.85             \\
                                                    &                                                                                                 & Yes                  & 67.08$\pm$1.94	& 60.17$\pm$1.09	& 49.08$\pm$1.99	& 47.13$\pm$1.56	& 39.76$\pm$1.43	& 52.64$\pm$9.96             \\ \cline{2-9} 
                                                    & IT                                                                              & No                   & 63.43$\pm$1.03	& 58.89$\pm$1.36	& 46.15$\pm$2.18	& 44.17$\pm$1.96	& 35.39$\pm$0.76	& 49.61$\pm$10.51             \\
                                                    & \begin{tabular}[c]{@{}l@{}}(continued pre-training) \\ (t5-base)\end{tabular}            & Yes                  & 65.85$\pm$0.78	& 60.97$\pm$0.69	& 49.82$\pm$1.10	& 47.38$\pm$1.23	& 39.20$\pm$1.17	& 52.64$\pm$9.71             \\ \cline{2-9} 
                                                    & IT-MTL                                                                        & No                   & 66.16$\pm$1.22	& 60.56$\pm$0.99	& 49.84$\pm$1.06	& 44.86$\pm$2.36	& 38.42$\pm$0.86	& 51.97$\pm$10.43             \\
                                                    & \begin{tabular}[c]{@{}l@{}}(continued pre-training)\\ (t5-base)\end{tabular}             & Yes                  & 68.00$\pm$1.07	& 61.54$\pm$1.14	& 50.66$\pm$1.09	& 48.11$\pm$1.08	& 40.35$\pm$1.30	& 53.73$\pm$9.97             \\ \hline
            \multirow{13}{*}{50}                    & \begin{tabular}[c]{@{}l@{}}Text \\ (t5-base)\end{tabular}                            & No                   & 69.06$\pm$0.70	& 63.97$\pm$0.59	& 55.42$\pm$0.70	& 50.50$\pm$0.99	& 45.91$\pm$1.56	& 56.97$\pm$8.73             \\ \cline{2-9} 
                                                    & \multirow{2}{*}{\begin{tabular}[c]{@{}l@{}}IT \\ (t5-base)\end{tabular}}        & No                   & 70.11$\pm$0.84	& 65.75$\pm$1.08	& 55.06$\pm$0.94	& 51.58$\pm$1.23	& 47.56$\pm$1.36	& 58.01$\pm$8.78             \\
                                                    &                                                                                                 & Yes                  & 70.14$\pm$0.97	& 65.13$\pm$0.82	& 55.86$\pm$0.95	& 52.63$\pm$0.94	& 47.53$\pm$1.02	& 58.26$\pm$8.36             \\ \cline{2-9} 
                                                    & \multirow{2}{*}{\begin{tabular}[c]{@{}l@{}}IT-MTL \\ (t5-base)\end{tabular}}  & No                   & 72.11$\pm$1.36	& 65.68$\pm$1.05	& 56.92$\pm$0.84	& 52.80$\pm$1.07	& 46.75$\pm$1.39	& 58.85$\pm$9.29             \\
                                                    &                                                                                                 & Yes                  & 71.92$\pm$0.88	& 65.88$\pm$0.70	& 56.56$\pm$0.99	& 53.83$\pm$1.08	& 47.88$\pm$1.37	& 59.21$\pm$8.72             \\ \cline{2-9} 
                                                    & \multirow{2}{*}{\begin{tabular}[c]{@{}l@{}}IT \\ (t5-large)\end{tabular}}       & No                   & 70.57$\pm$0.96	& 67.34$\pm$1.68	& 58.99$\pm$1.29	& 53.13$\pm$0.93	& 48.87$\pm$0.94	& 59.78$\pm$8.46             \\
                                                    &                                                                                                 & Yes                  & 71.77$\pm$0.77	& 66.66$\pm$1.11	& 59.59$\pm$1.44	& 55.06$\pm$1.45	& 50.36$\pm$0.89	& 60.69$\pm$7.88             \\ \cline{2-9} 
                                                    & \multirow{2}{*}{\begin{tabular}[c]{@{}l@{}}IT-MTL \\ (t5-large)\end{tabular}} & No                   & 71.73$\pm$0.55	& 66.65$\pm$1.05	& 57.89$\pm$0.76	& 53.17$\pm$2.33	& 47.69$\pm$1.62	& 59.42$\pm$9.02             \\
                                                    &                                                                                                 & Yes                  & 72.38$\pm$0.83	& 66.70$\pm$0.77	& 58.48$\pm$1.27	& 53.89$\pm$1.56	& 48.45$\pm$1.53	& 59.98$\pm$8.78             \\ \cline{2-9} 
                                                    & IT                                                                              & No                   & 69.80$\pm$1.11	& 65.11$\pm$0.51	& 55.94$\pm$1.51	& 50.75$\pm$1.06	& 45.25$\pm$1.11	& 57.37$\pm$9.27             \\
                                                    & \begin{tabular}[c]{@{}l@{}}(continued pre-training) \\ (t5-base)\end{tabular}            & Yes                  & 70.06$\pm$1.29	& 64.81$\pm$1.12	& 55.68$\pm$0.95	& 52.12$\pm$0.98	& 46.69$\pm$1.49	& 57.87$\pm$8.62             \\ \cline{2-9} 
                                                    & IT-MTL                                                                        & No                   & 72.08$\pm$0.79	& 66.74$\pm$0.99	& 58.02$\pm$0.95	& 52.48$\pm$1.77	& 46.66$\pm$1.35	& 59.19$\pm$9.49             \\
                                                    & \begin{tabular}[c]{@{}l@{}}(continued pre-training)\\ (t5-base)\end{tabular}             & Yes                  & 71.20$\pm$0.87	& 65.79$\pm$1.19	& 56.68$\pm$0.96	& 53.31$\pm$0.87	& 47.10$\pm$0.85	& 58.82$\pm$8.76             \\ \hline
        \end{tabular}
    }
    \caption{\footnotesize{Comparison of $k$-Shot performances on all $5$ ABSA tasks for \textsc{Rest16}.}}
    \label{appendix_ss:kshot_rest16}
\end{table*}

\subsection{Cross Domain}
\label{appendix_ss:cross_domain}
We experiment with pre-training on a different domain than the domain of the downstream task. 
Concretely, we perform two experiments: (i) we perform NAPT on restaurant domain, then finetune on the laptop domain, and (ii) we perform NAPT on the laptop domain, then finetune on the restaurant domain. 
We include the results with our proposed model trained with NAPT on restaurant data and finetuned on \textsc{Lap14} in Table~\ref{appendix_ss:cross_domain_lap14}.
We include the results with our proposed model trained with NAPT on laptop data and finetuned on 
\textsc{Rest15} and \textsc{Rest16} in Table~\ref{appendix_ss:cross_domain_rest15} and in Table~\ref{appendix_ss:cross_domain_rest16}, respectively.
We observed that our proposed model is still able to transfer the knowledge learned during the NAPT phase. 
Our proposed model still outperforms the baseline, brining as much as $11.49\%$~F1 points for the ASTE task in the laptop domain.
In some cases we noticed a slight increase in the final performance compared to the model trained with NAPT on the full dataset. 
This suggests that the model trained on the full dataset overfits to the noisy data.

\begin{table*}[h]
    \centering
    \resizebox{0.5\textwidth}{!}{
        \begin{tabular}{c|ccc|c}
        \hline
            \multicolumn{1}{c|}{\textbf{k}} & \multicolumn{1}{c}{\textbf{AE}} & \multicolumn{1}{c}{\textbf{AESC}} & \multicolumn{1}{c|}{\textbf{ASTE}} & \multicolumn{1}{c}{\textbf{Average}} \\ \hline
            5                               & 47.55$\pm$2.06                      & 36.55$\pm$2.35                        & 24.53$\pm$2.25                         & 33.06                                \\
            10                              & 55.93$\pm$2.80                      & 45.55$\pm$2.39                        & 35.38$\pm$1.80                         & 43.33                                \\
            20                              & 64.55$\pm$1.47                      & 52.18$\pm$1.07                        & 41.67$\pm$1.97                         & 52.51                                \\
            50                              & 69.52$\pm$0.71                      & 56.25$\pm$1.44                        & 46.49$\pm$1.97                         & 57.30                                \\
            Full Dataset                    & 77.32$\pm$1.18                      & 68.20$\pm$0.72                        & 60.93$\pm$1.12                         & 68.56     \\ \hline                          
        \end{tabular}
    }
    \caption{\footnotesize{Cross-Domain performance of {\tt IT-MTL-NAPT} on \textsc{Lap14}. The NAPT was done only on \textit{Yelp} corpus.}}
    \label{appendix_ss:cross_domain_lap14}
\end{table*}

\begin{table*}[h]
    \centering
    \resizebox{0.7\textwidth}{!}{
        \begin{tabular}{c|ccccc|c}
        \hline
            \multicolumn{1}{c|}{\textbf{k}} & \multicolumn{1}{c}{\textbf{AE}} & \multicolumn{1}{c}{\textbf{AESC}} & \multicolumn{1}{l}{\textbf{TASD}} & \multicolumn{1}{c}{\textbf{ASTE}} & \multicolumn{1}{l|}{\textbf{ASQP}} & \multicolumn{1}{c}{\textbf{Average}} \\ \hline
            5                               & 53.17$\pm$2.79                      & 44.54$\pm$1.97                        & 29.26$\pm$1.96                        & 32.89$\pm$1.58               & 21.75$\pm$1.25                         & 35.80                                \\
            10                              & 63.07$\pm$1.43                      & 53.79$\pm$2.13                        & 38.05$\pm$1.82                        & 42.22$\pm$1.76               & 29.80$\pm$2.15                         & 45.41                                \\
            20                              & 68.99$\pm$1.34                      & 60.20$\pm$1.21                        & 44.84$\pm$1.23                        & 46.01$\pm$1.22               & 35.18$\pm$1.55                         & 52.02                                \\
            50                              & 74.20$\pm$0.89                      & 64.50$\pm$0.85                        & 50.67$\pm$1.08                        & 50.18$\pm$1.65               & 40.49$\pm$1.37                         & 57.54                                \\
            Full Dataset                    & 79.39$\pm$1.07                      & 72.37$\pm$1.02                        & 62.92$\pm$1.11                        & 58.95$\pm$1.11               & 51.38$\pm$0.90                         & 65.32 \\ \hline                              
        \end{tabular}
    }
    \caption{\footnotesize{Cross-Domain performance of {\tt IT-MTL-NAPT} on \textsc{Rest15}. The NAPT was done only on \textit{Amazon Reviews} corpus.}}
    \label{appendix_ss:cross_domain_rest15}
\end{table*}

\begin{table*}[h]
    \centering
    \resizebox{0.7\textwidth}{!}{
        \begin{tabular}{c|ccccc|c}
        \hline
            \multicolumn{1}{c|}{\textbf{k}} & \multicolumn{1}{c}{\textbf{AE}} & \multicolumn{1}{c}{\textbf{AESC}} & \multicolumn{1}{l}{\textbf{TASD}} & \multicolumn{1}{c}{\textbf{ASTE}} & \multicolumn{1}{l|}{\textbf{ASQP}} & \multicolumn{1}{c}{\textbf{Average}} \\ \hline
            5                               & 59.17$\pm$1.63                      & 54.07$\pm$1.35                        & 38.05$\pm$2.04                        & 41.03$\pm$1.68               & 29.26$\pm$1.74                         & 43.46                                \\
            10                              & 62.80$\pm$1.54                      & 57.27$\pm$1.71                        & 42.65$\pm$2.11                        & 43.66$\pm$1.44               & 34.14$\pm$1.18                         & 47.74                                \\
            20                              & 66.06$\pm$1.21                      & 60.46$\pm$1.58                        & 47.96$\pm$1.34                        & 47.10$\pm$1.30               & 38.32$\pm$1.02                         & 52.31                                \\
            50                              & 69.67$\pm$1.12                      & 64.61$\pm$0.76                        & 54.17$\pm$1.40                        & 51.91$\pm$1.08               & 45.29$\pm$1.25                         & 57.80                                \\
            Full Dataset                    & 80.72$\pm$0.81                      & 75.72$\pm$0.89                        & 68.95$\pm$0.97                        & 64.04$\pm$0.84               & 58.02$\pm$0.97                         & 68.84 \\ \hline                               
        \end{tabular}
    }
    \caption{\footnotesize{Cross-Domain performance of {\tt IT-MTL-NAPT} on \textsc{Rest16}. The NAPT was done only on \textit{Amazon Reviews} corpus.}}
    \label{appendix_ss:cross_domain_rest16}
\end{table*}
\section{Multi-word Patterns}
\label{appendix:multiword_patterns}

In Table \ref{appendix_ss:multiword_patterns} we list all the patterns that were used to filter 2-grams, 3-grams and 4-grams during aspect term extraction.

\begin{table*}[]
\centering
    \resizebox{0.2\textwidth}{!}{
\begin{tabular}{|c|}
\hline
\textbf{Multi-word Patterns}        \\ \hline
NN*-NN*         \\ \hline
JJ*-NN*         \\ \hline
VBG-NN*         \\ \hline
VBN-NN*         \\ \hline
NN*-NN*-NN*     \\ \hline
NN*-IN-NN*      \\ \hline
JJ*-NN*-NN*     \\ \hline
JJ*-JJ*-NN*     \\ \hline
VBN-JJ*-NN*     \\ \hline
NN*-NN*-NN*-NN* \\ \hline
NN*-CC-NN*-NN*  \\ \hline
\end{tabular}
}
\caption{Multi-word Patterns used to filter 2-grams, 3-grams and 4-grams. `*' denotes any variant of the corresponding POS tags. For example, NN* captures NN, NNS, NNP, NNPS.}
\label{appendix_ss:multiword_patterns}
\end{table*}

\end{document}